\documentclass[runningheads]{llncs}
\usepackage{graphicx}
\usepackage{amsmath,amssymb} 
\usepackage{color}
\usepackage{url}
\usepackage[font=small,skip=1pt]{caption}
\usepackage[font=small]{subfig}
\usepackage{comment}
\usepackage{multirow}
\usepackage{booktabs}
\usepackage{xcolor}
\usepackage{tabularx}
\usepackage[super]{nth}
\newcolumntype{Y}{>{\centering\arraybackslash}X}
\usepackage{enumitem}
\usepackage{algorithm}
\usepackage{algorithmic}
\usepackage{color,soul}
\usepackage{bbding}
\soulregister\cite7
\soulregister\ref7
\soulregister\pageref7
\newcommand{\bfx}{{\bf x}}
\newcommand{\bfI}{{\bf I}}
\newcommand{\bfJ}{{\bf J}}
\newcommand{\bfA}{{\bf A}}

\begin{document}
\pagestyle{headings}
\mainmatter

\def\ACCV22SubNumber{393}  

\title{Structure Representation Network and Uncertainty Feedback Learning for \\Dense Non-Uniform Fog Removal} 
\titlerunning{Structure Representation and Uncertainty Feedback for Fog Removal}

\author{Yeying Jin\inst{1}\orcidID{0000-0001-7818-9534} \and
	Wending Yan\inst{1,3}\orcidID{0000-0001-5993-8405} \and
	Wenhan Yang\inst{2}\orcidID{0000-0002-1692-0069} \and
	Robby T. Tan\inst{1,3}\orcidID{0000-0001-7532-6919}}
\institute{National University of Singapore, \and Nanyang Technological University,\and Yale-NUS College\\
{\tt\small 
jinyeying@u.nus.edu, e0267911@u.nus.edu, wenhan.yang@ntu.edu.sg, robby.tan@\{nus,yale-nus\}.edu.sg}}

\authorrunning{Y. Jin, W. Yan et al.}
\def\thefootnote{$\dagger$}\footnotetext{Our data and code is available at: \url{https://github.com/jinyeying/FogRemoval}}\def\thefootnote{\arabic{footnote}}

\maketitle

\begin{abstract}
Few existing image defogging or dehazing methods consider dense and non-uniform particle distributions, which usually happen in smoke, dust and fog.
Dealing with these dense and/or non-uniform distributions can be intractable, since fog's attenuation and airlight (or veiling effect) significantly weaken the background scene information in the input image.
To address this problem, we introduce a structure-representation network with uncertainty feedback learning.
Specifically, we extract the feature representations from a  pre-trained Vision Transformer (DINO-ViT) module to recover the background information.
To guide our network to focus on non-uniform fog areas, and then remove the fog accordingly, we introduce the uncertainty feedback learning, which produces uncertainty maps, that have higher uncertainty in denser fog regions, and can be regarded as an attention map that represents fog's density and uneven distribution.
Based on the uncertainty map, our feedback network refines our defogged output iteratively.
Moreover, to handle the intractability of estimating the atmospheric light colors, we exploit the grayscale version of our input image, since it is less affected by varying light colors that are possibly present in the input image.
The experimental results demonstrate the effectiveness of our method both quantitatively and qualitatively compared to the state-of-the-art methods in handling dense and non-uniform fog or smoke.
\end{abstract}

\begin{figure}[t]
	\centering
		\captionsetup[subfloat]{labelformat=empty}
		\captionsetup[subfloat]{farskip=2pt}
		\setcounter{subfigure}{0}
		\subfloat[Input]{\includegraphics[width=0.245\textwidth]{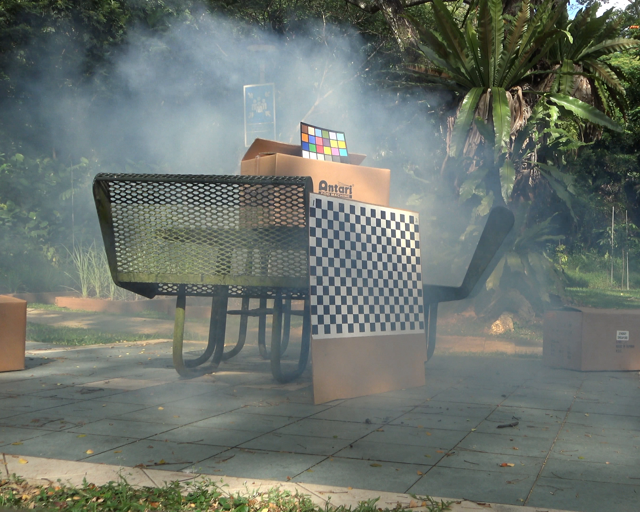}}\hfill
		\subfloat[Ours]{\includegraphics[width=0.245\textwidth]{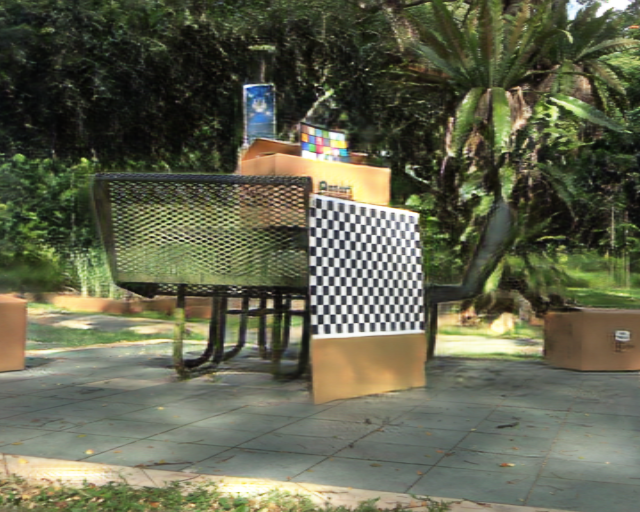}}\hfill
		\subfloat[D4'22~\cite{yang2022self}]{\includegraphics[width=0.245\textwidth]{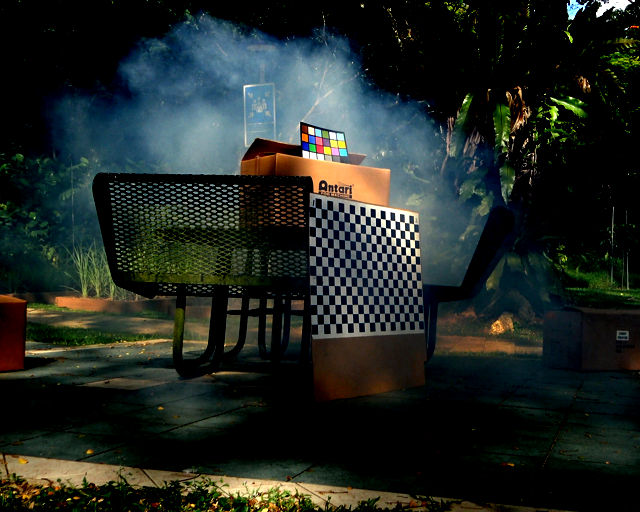}}\hfill
		\subfloat[DeHamer'22~\cite{guo2022image}]{\includegraphics[width=0.245\textwidth]{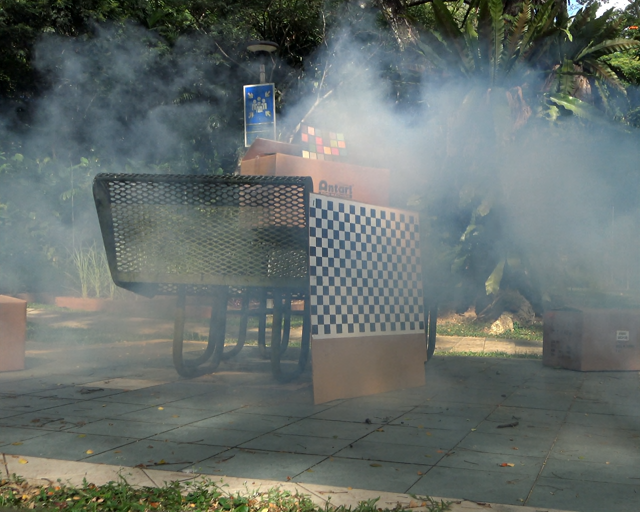}}\hfill
	\caption{Visual comparisons of different methods: the state-of-the-art CNN-based method~\cite{yang2022self} and transformer-based method~\cite{guo2022image} in dense and/or non-uniform fog.}
	\label{fig:intro}
\end{figure}

\section{Introduction}
\label{sec:intro}
Atmospheric particles, such as fog, haze, dust and smoke particles, can degrade the visibility of a scene significantly as shown in Fig.~\ref{fig:intro}.
These particles can be modeled as~\cite{koschmieder1924theorie}:
\begin{equation}
	\bfI(\bfx) = \bfJ(\bfx)t(\bfx) + \left(1-t(\bfx)\right)\bfA,
	\label{eq:asmodel}
\end{equation}
where $\bfI$ is an observed RGB color vector, $\mathbf{x}$ is the pixel location. $\bfJ$ is the scene radiance. $\bfA$ is the atmospheric light, and $t$ is the transmission. 
The first term is called direct attenuation, and the second term is called airlight. 
Transmission $t$ can be modeled as $t(\mathbf{x}) = \exp(\beta(\mathbf{x}) d(\mathbf{x}))$, 
where $\beta$ is the particle attenuation factor that depends on the density of the particle distribution and the size of particles;
while $d$ is the depth of the scene with respect to the camera. 
Most existing methods assume the uniformity of the particle distributions, which means they assume $\beta$ to be independent from $\mathbf{x}$.
Note that, in this paper, we deal with fog, haze, atmospheric dust and smoke that can be dense and/or non-uniform. However, for clarity, we write fog to represent them.

Many methods have been proposed to deal with fog degradation.
Existing fully supervised CNN-based methods~\cite{cai2016dehazenet,li2017aod,zhang2018densely,qu2019enhanced,liu2019griddehazenet,dong2020multi,wu2021contrastive} require clean ground truths, which are intractable to obtain particularly for non-uniform fog.
Synthetic images, unfortunately, cannot help that much for dense and/or non-uniform fog. Synthesizing non-uniform fog is difficult and computationally expensive, and dense synthetic fog has significant gaps with real dense fog.
Semi-supervised methods~\cite{li2019semi,shao2020domain,chen2021psd,li2022physically} adopt the domain adaptation. However, the huge domain gap between synthetic and real dense and/or non-uniform fog images is not easy to align.
Unsupervised methods~\cite{huang2019towards,golts2019unsupervised,li2021you,zhao2021refinednet,yang2022self} make use of statistical similarity between unpaired training data,  and are still less effective compared with semi-supervised or supervised methods. Importantly, unsupervised methods can generate hallucinations, particularly in dense fog areas.
Recently, ViT-based dehazing methods~\cite{guo2022image,song2022vision} have been proposed; however, memory and computation complexity slow down the convergence~\cite{zhu2020deformable}, causing unreliable performance on real-world high-resolution non-uniform fog images.

In this paper, our goal is to remove fog, particularly dense or non-uniform fog, or a combination of the two (dense and non-uniform).
Unlike non-dense uniform fog, where the human visual perception can still discern the background scenes, dense and/or non-uniform fog significantly weakens the information of the background scenes (see Fig.~\ref{fig:intro}).
To achieve our goal, first, we exploit the representation extracted from DINO-ViT~\cite{caron2021emerging}, a self-supervised pre-trained model in order to recover background structures.
DINO-ViT captures visual representations from data, e.g., scene structure representations, based on self-similarity prior~\cite{shechtman2007matching}.
Second, since the recovery of the $\mathbf{A}$ is challenging~\cite{Sulami2014}, to avoid the direct recovery of $\mathbf{A}$, we introduce a grayscale feature multiplier to learn fog degradation in an end-to-end manner.
Grayscale images are less affected by multi-colored light sources (skylight, sunlight, or cars' headlights, etc.) as well as the colors of the particle scattered lights (whitish for fog, yellowish or reddish for haze or atmospheric dust or smoke).
We can multiply the grayscale features and our feature multiplier (derived from the model in Eq.~\eqref{eq:asmodel}), to ensure our features are unaffected by the airlight and thus are more reliable when applied to our multiplier consistency loss.

Third, we propose an uncertainty-based feedback learning that allows our network to pay more attention to regions that are still affected by fog based on our uncertainty predictions, iteratively.
Since the network usually has high uncertainty on the dense fog regions (because background information is washed out by the fog and the input image contains less background information in those regions), we can use an uncertainty map as an attention cue to guide the network to differentiate dense fog region from the rest of input image.
In one iteration, if our output still contains fog in some regions, the uncertainty map will indicate those regions, and in the next iteration, our method will focus on these regions to further defog them.

To sum up, our main contributions and novelties are as follows:
\begin{itemize}[noitemsep,topsep=1pt]
	\item 
	To the best of our knowledge, our method is the first single-image defogging network that performs robustly in dense non-uniform fog, by combining structure representations from ViT and features from CNN as feature regularization. Thus, the background information under fog can be preserved and extracted.
	\item
	We propose the grayscale feature multiplier that acts as feature enhancement and guides our network to learn to extract clear background information.
	\item 
	We introduce the uncertainty feedback learning in our defogging network, which can refine the defogging results iteratively by focusing on areas that still suffer from fog.
\end{itemize}
Experimental results show that our method is effective in removing dense and/or non-uniform fog images, outperforming the state-of-the-art methods both quantitatively and qualitatively.

\section{Related Works}
\label{sec:related}
Non-learning methods introduced priors from the atmosphere scattering model.
Tan~\cite{tan2008visibility} estimates the airlight to increase contrast, 
Fattal~\cite{fattal2008single} estimates transmission, which is statistical uncorrelated to surface shading,
He et al.~\cite{he2010single} introduce the dark channel prior, 
Berman et al.~\cite{berman2018single} propose a haze-line constraint,
and Meng et al.~\cite{meng2013efficient} estimate transmission using its minimum boundary. 

CNN-based methods allow faster results~\cite{li2017haze,ye2021perceiving,lin2020gait,lin2021gait}.
DehazeNet~\cite{cai2016dehazenet} and MSCNN~\cite{ren2016single} use CNN, DCPDN~\cite{zhang2018densely} trains densely connected pyramid network to estimate transmission map.
AODNet~\cite{li2017aod}, GFN~\cite{ren2018gated},~\cite{li2018single} applies CGAN, EPDN~\cite{qu2019enhanced} applies pix2pix, they end-to-end output clear images.
Griddehazenet~\cite{liu2019griddehazenet} designs attention-based~\cite{jin2021dc} grid network, MSBDN~\cite{dong2020multi} designs boosted multi-scale decoder, FFA-Net~\cite{qin2020ffa} proposes feature fusion attention network, AECR-Net~\cite{wu2021contrastive} applies contrastive learning. 
Few fully-supervised methods~\cite{dudhane2019ri,bianco2019high,morales2019feature} are proposed to deal with Dense-Haze~\cite{ancuti2019dense}.
All these methods employ fully supervised learning and hence require ground truths to train their networks. 
However, obtaining a large number of real dense or non-uniform fog images and their corresponding ground truths is intractable. 
Semi-supervised methods~\cite{li2019semi,shao2020domain,chen2021psd,li2022physically} have been introduced, 
unfortunately, they still suffer from gaps between synthetic and real fog images.
Unsupervised methods~\cite{huang2019towards,golts2019unsupervised,li2021you,zhao2021refinednet,yang2022self} are mainly CycleGAN-based.
However, the generated images can easily render artefacts (structures that are not originally in the input image) when unpaired training data is used.
Though all these methods perform well on normal fog dataset, they are CNN-based, and tend to perform poorly on dense and non-uniform fog~\cite{guo2022image} since CNN fails to model long-range pixel dependencies~\cite{dosovitskiy2020image}.
 
Recently, ViT-based dehazing~\cite{guo2022image,song2022vision} has made progress.
DehazeFormer~\cite{song2022vision} is trained on synthetic fog images (RESIDE outdoor dataset~\cite{li2018benchmarking}), which are not realistic and cause unreliable performance on real-world fog images.
DeHamer~\cite{guo2022image} combines CNN and Transformer for image dehazing; however, memory and computation complexity slow down the convergence~\cite{zhu2020deformable}, causing inefficient performance on real-world high resolution fog images.
In contrast, our method exploits features from both ViT and CNN.

\begin{figure*}[t!]
	\centering
	{\includegraphics[width=1\textwidth]{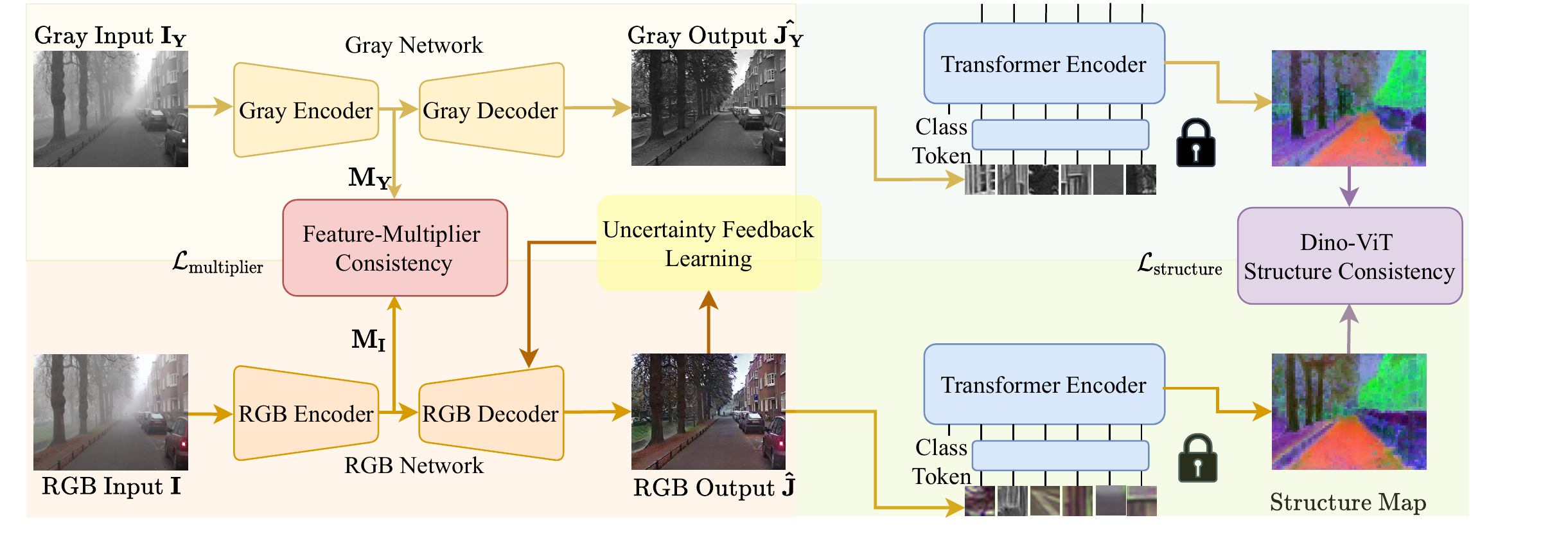}}\hfill
	\caption{The pipeline of our network, which consists of (i) grayscale feature multiplier (top left), (ii) structure representation network (right), and (iii) uncertainty feedback learning (middle).
	The grayscale feature multiplier ($\mathbf{M_Y}$) provides features (red) from CNN, and guides the RGB network to enhance features. 
	The structure representation network provides structure representations (purple) from fixed and pre-trained DINO-ViT, to recover background information.}
	\label{fig:pipeline}
\end{figure*}

\section{Proposed Method}\label{sec:proposed}
Fig.~\ref{fig:pipeline} shows the pipeline of our architecture, which consists of three parts: (i) grayscale feature multiplier, (ii)  structure representation network, and (iii) uncertainty feedback learning.
Since grayscale images are less affected by multi-colored light sources and colorful particle scattered lights, we develop a grayscale network to guide our RGB network.
Hence, in our pipeline, we have two parallel subnetworks: one for processing the grayscale input image,  and the other one for processing the RGB input image.

\subsection{Grayscale Feature Multiplier}
\begin{figure}[t]
	\centering
	\subfloat[Input $\mathbf{I}$]{\includegraphics[width = 0.196\columnwidth]{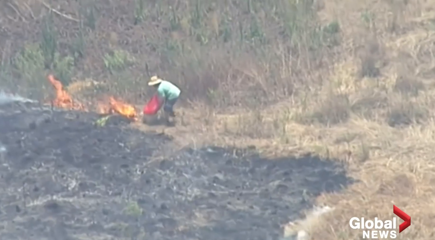}}\hfill
	\subfloat[Gray $\mathbf{\hat{J}_Y}$]{\includegraphics[width = 0.196\columnwidth]{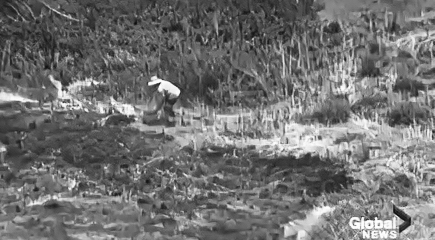}}\hfill
	\subfloat[$\mathbf{M(\mathbf{\hat{J}_Y})}$]{\includegraphics[width = 0.196\columnwidth]{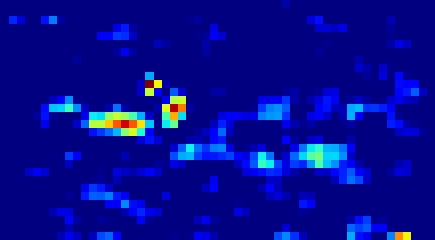}}\hfill
	\subfloat[Output $\mathbf{\hat{J}}$]{\includegraphics[width = 0.196\columnwidth]{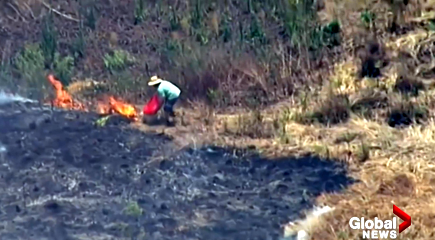}}\hfill
	\subfloat[$\mathbf{M(\mathbf{\hat{J}})}$]{\includegraphics[width = 0.196\columnwidth]{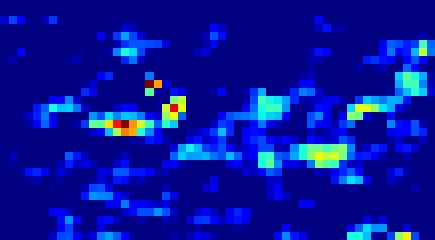}}\hfill\\
	\caption{Visualization of the features extracted from the grayscale feature multiplier.
		(a) Input fog image $\mathbf{I}$, 
		(b) Grayscale output image $\mathbf{\hat{J}_Y}$,
		(c) Sample feature map for $\mathbf{\hat{J}_Y}$, 
		(d) Output fog-free image  $\mathbf{\hat{J}}$, 
		and (e) Sample feature map for  $\mathbf{\hat{J}}$. 
		We can observe that features in (c) for the grayscale fog images are less affected by fog, and can effectively guide the features in (e) owing to our multiplier consistency loss.}
	\label{fig:featureloss}
\end{figure}

\subsubsection{Feature Multiplier}
Dense and/or non-uniform fog suffers from low contrast and degraded features. 
To extract clear background features, we design a subnetwork to predict the amount by which these features should be enhanced.
Considering the fog model in Eq.~\eqref{eq:asmodel} and to avoid the challenges of predicting atmosphere light $\mathbf{A}$~\cite{Sulami2014}, 
we turn the relationship between fog $\mathbf{I}$ and clear images $\mathbf{J}$ into a multiplier relationship:
$\bfJ(\bfx) = \bfI(\bfx) \bf{M}(\bfx)$, which is called $\bf{M}$ feature multiplier~\cite{li2019rainflow}, where 
$\mathbf{M}(\bfx) =  \frac{\bfI(\bfx)+t(\bfx)\bfA-\bfA}{\bfI(\bfx)t(\bfx)}$.

The feature multiplier $\mathbf{M}$ depends on atmospheric light $\mathbf{A}$ and transmission $t(\mathbf{x})$, which are both unknown. 
Moreover, $\mathbf{A}$ is an RGB color vector; implying that in order to estimate $\mathbf{M}$, there are four unknowns in total for each pixel: 3 for the RGB values of $\mathbf{A}$ and 1 for $t$. 
These unknowns influence the accuracy of the network in learning the correct value of $\mathbf{M}$. 
To overcome the difficulty, we propose to employ a grayscale feature multiplier, where all variables in the grayscale feature multiplier become scalar variables. 
Consequently, the number of unknowns the network needs to learn is reduced to only two variables for each pixel: $t(\mathbf{x})$ and $\mathbf{A}$.
Note that, to avoid the direct recovery of $\mathbf{A}$, our network implicitly includes $\mathbf{A}$ in the feature multiplier.
 
\subsubsection{Grayscale-Feature Multiplier}
We feed the grayscale image, $\mathbf{I_Y}$, to our grayscale encoder, which estimates the grayscale feature multiplier $\mathbf{M_Y}$. 
We multiply grayscale features and $\mathbf{M_Y}$ before feeding them to our grayscale decoder.
We train the grayscale network independently from the RGB network, using both synthetic and unpaired real images.
Once the grayscale network is trained, we freeze it, and employ it as the guidance for training the RGB network.

As for the RGB network, the RGB encoder takes the RGB image as input, $\mathbf{I}$, and estimates the color feature multiplier $\mathbf{M_I}$. 
Having estimated $\mathbf{M_I}$, we multiply it with the RGB features and feed the multiplied features to our RGB decoder.
As shown in Fig.~\ref{fig:pipeline}, we constrain the learning process of our RGB network by imposing a consistency loss between the grayscale feature multiplier, $\mathbf{M_Y}$, and the RGB feature multiplier $\mathbf{M_I}$.
We call this loss a multiplier consistency loss.

\subsubsection{Multiplier Consistency Loss} 
To constrain the RGB feature multiplier $\mathbf{M_I}$, we utilize the grayscale feature multiplier $\mathbf{M_Y}$ as guidance.
Based on the Gray World assumption~\cite{buchsbaum1980spatial},  we define the multiplier consistency loss as:
\begin{align}
\mathcal{L}_{\rm multiplier} = \left \| \mathbf{M_I}-\mathbf{{M_Y}} \right \| _{2},
\label{eq:loss_multi_real}
\end{align} 
where $\mathbf{M_I}$ and $\mathbf{M_Y}$ are the feature multipliers of the RGB and grayscale images.
To construct this loss, first, we train our grayscale network independently from our RGB network.
By training the grayscale network on both synthetic and real images, $\mathbf{M_Y}$ is optimized. 
Once the training is completed, we freeze the grayscale network.
Subsequently, we train our RGB network. 

In this training stage, the network losses are the same as those in the grayscale network, except all the images used to calculate the losses are now RGB images.  
Unlike the training process of the grayscale network, however, we need to apply the multiplier consistency loss $\mathcal{L}_{\rm multiplier}$ to train the RGB network.
Note that, the reason we use the loss to enforce $\mathbf{M_I}$ and $\mathbf{M_Y}$ to be close, and do not use $\mathbf{M_Y}$ as the feature multiplier for the RGB network (i.e., $\mathbf{M_I} = \mathbf{M_Y}$) is  because we intend to train the RGB convolution layers; 
so that, in the testing stage, we do not need the grayscale network. 

\begin{figure}[t]
	\centering
	\setcounter{subfigure}{0}
	\subfloat[Input $\mathbf{I}$]{\includegraphics[width=0.196\textwidth]{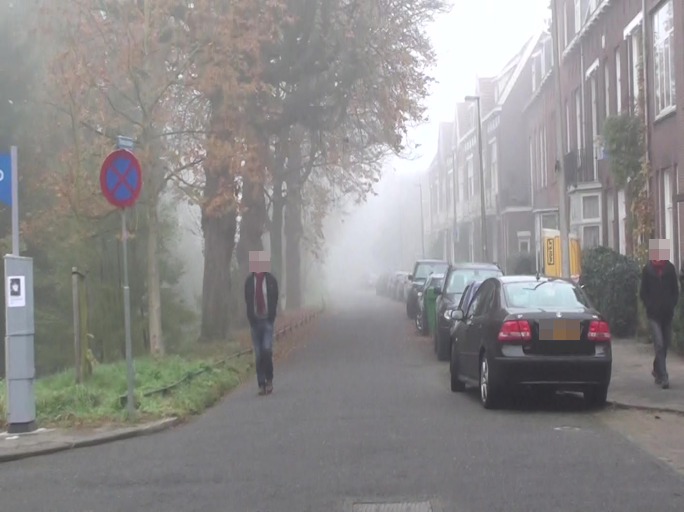}}\hfill
	\subfloat[Gray $\mathbf{\hat{J}_Y}$]{\includegraphics[width=0.196\textwidth]{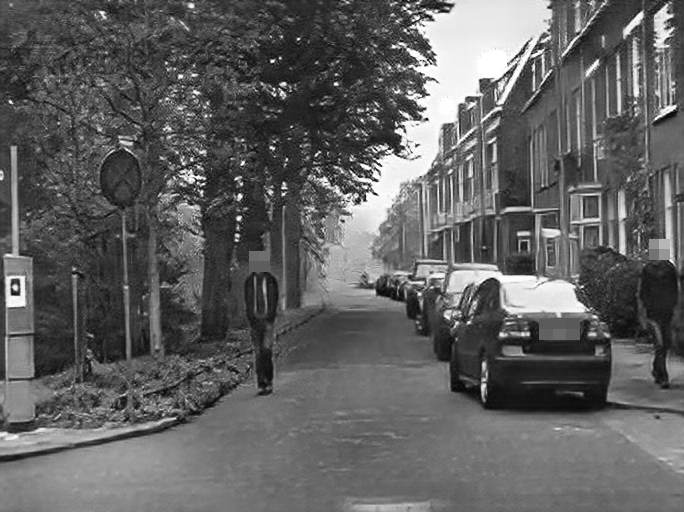}}\hfill
	\subfloat[$S(\mathbf{\hat{J}_Y})$]{\includegraphics[width=0.196\textwidth]{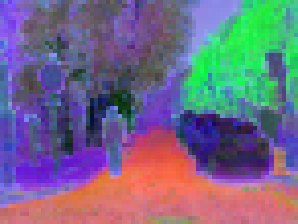}}\hfill
	\subfloat[Output $\mathbf{\hat{J}}$]{\includegraphics[width=0.196\textwidth]{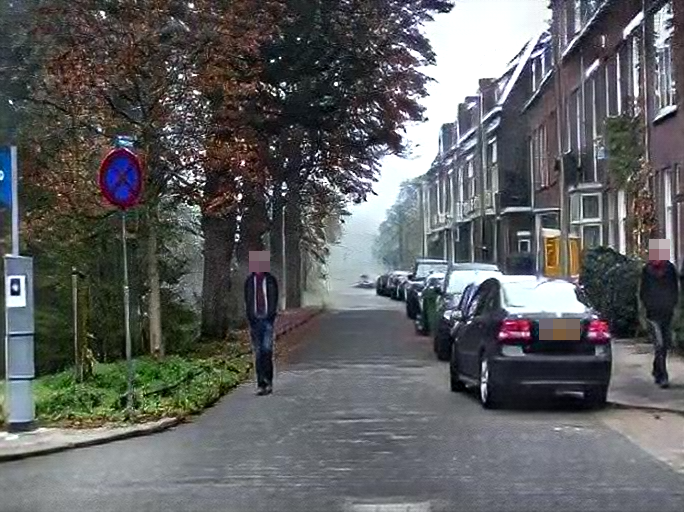}}\hfill
	\subfloat[$S(\mathbf{\hat{J}})$]{\includegraphics[width=0.196\textwidth]{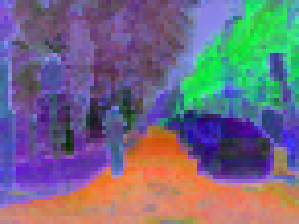}}\hfill
	\caption{Visualization of structure representations.
		(a) Input fog image $\mathbf{I}$, 
		(b) Grayscale output image $\mathbf{\hat{J}_Y}$,
		(c) DINO-ViT keys for $\mathbf{\hat{J}_Y}$, 
		(d) Output fog-free image $\mathbf{\hat{J}}$, 
		and (e) DINO-ViT keys for $\mathbf{\hat{J}}$. 
		We can observe that DINO-ViT representations in (c) capture structure scene/object parts (\textit{e.g.} cars, trees, buildings), and are less affected by fog.}
	\label{fig:structure}
\end{figure}

\subsection{Structure Representation Network}
A few methods ~\cite{shechtman2007matching,zheng2021spatially,kolkin2019style,jin2022unsupervised} have exploited self-similarity-based feature descriptors to obtain structure representations. 
Unlike these methods, to reveal the clear background structures, we use deep spatial features obtained from DINO-ViT~\cite{tumanyan2022splicing}, which has been proven to learn meaningful visual representations~\cite{amir2021deep}. Moreover, these powerful representations are shared across different object classes.
Specifically, we use keys' self-similarity in the attention model, at the deepest transformer layer.
In Fig.~\ref{fig:structure}, we show the Principal Component Analysis (PCA) visualization of the keys' self-similarity and demonstrate the three top components as RGB at layer 11 of DINO-ViT. 
As one can observe, the structure representations capture the clear background parts, which helps the network significantly preserve the background structures.  

\subsubsection{Dino-ViT Structure Consistency Loss}
Our Dino-ViT structure consistency loss encourages the deep-structure representations of the RGB output to be similar to the grayscale features, since the grayscale features are robust to fog: 
\begin{align}
\mathcal{L}_{\rm structure} = \left \|  S(\mathbf{\hat{J}})-S(\mathbf{\hat{J}_Y}) \right \| _{F},
\label{eq:loss_structure}
\end{align} 
where $S$ is the self-similarity descriptor, defined by the difference in the self-similarity of the keys extracted from the attention module, with $n \times n$ dimension, where $n$ is the number of patches. 
$\left \| \cdot \right \| _{F}$ is the Frobenius norm.
The self-similarity descriptor is defined as:
\begin{align}
S(\mathbf{\hat{J}})_{ij} = \text{cos-sim}(k_i(\mathbf{\hat{J}}), k_j(\mathbf{\hat{J}})) = 1 - \frac{k_i(\mathbf{\hat{J}}) \cdot k_j(\mathbf{\hat{J}})} {\left \| k_i(\mathbf{\hat{J}}) \right \| \cdot \left \| k_j(\mathbf{\hat{J}}) \right \| },
\label{eq:loss_sel}
\end{align} 
where $\text{cos-sim}(\cdot)$ is the cosine similarity between keys, $k_i$ are the spatial keys. 

\begin{figure}[t]
	\centering
		\captionsetup[subfloat]{labelformat=empty}
		\setcounter{subfigure}{0}
		\subfloat[Input $\mathbf{I}$]{\includegraphics[width=0.163\textwidth,height=0.17\textwidth]{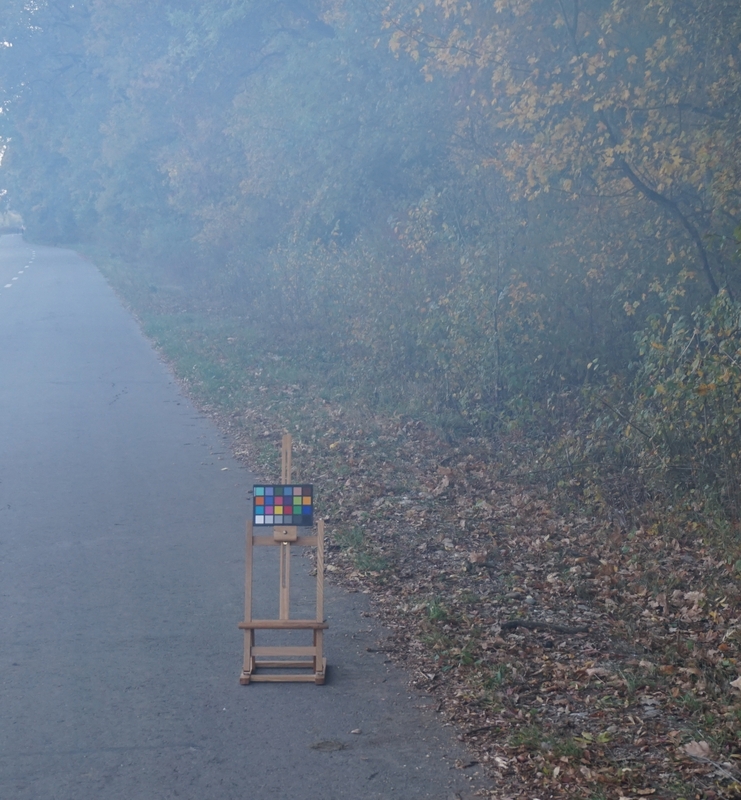}}\hfill
		\subfloat[Uncertainty $\theta$]{\includegraphics[width=0.163\textwidth,height=0.17\textwidth]{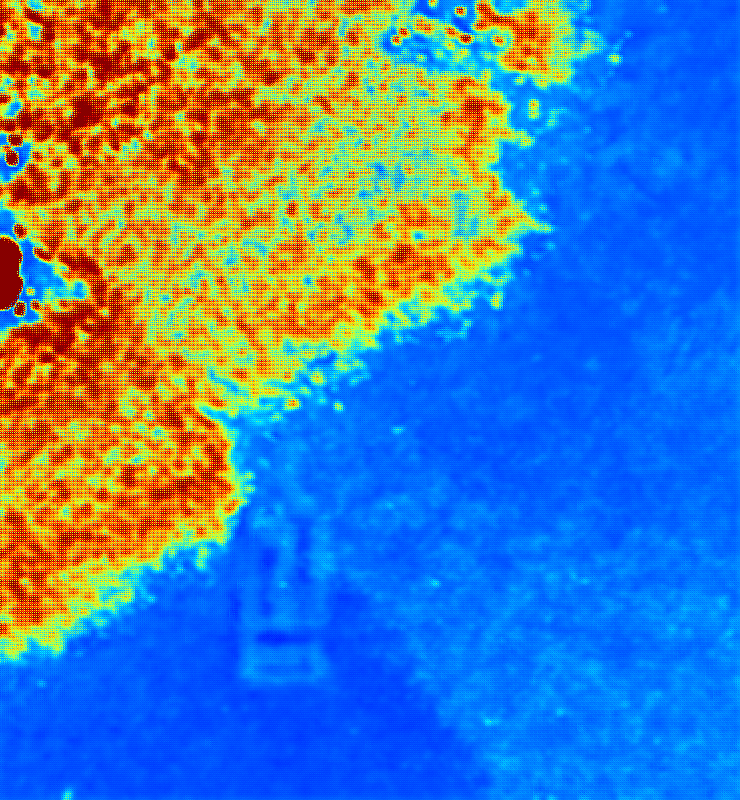}}\hfill
		\subfloat[Output]{\includegraphics[width=0.163\textwidth,height=0.17\textwidth]{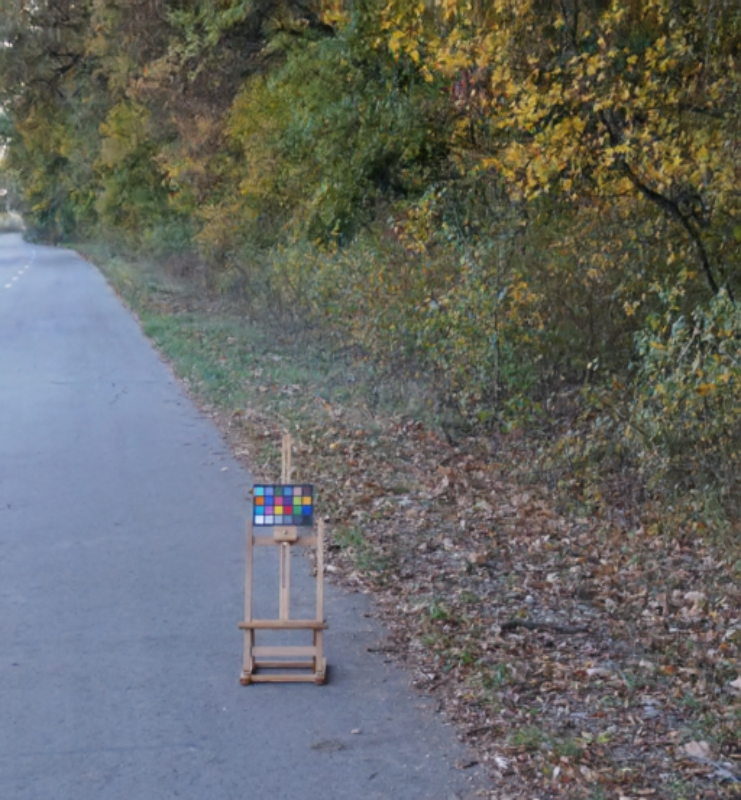}}\hfill
		\setcounter{subfigure}{0}
		\subfloat[Input $\mathbf{I}$]{\includegraphics[width=0.163\textwidth,height=0.17\textwidth]{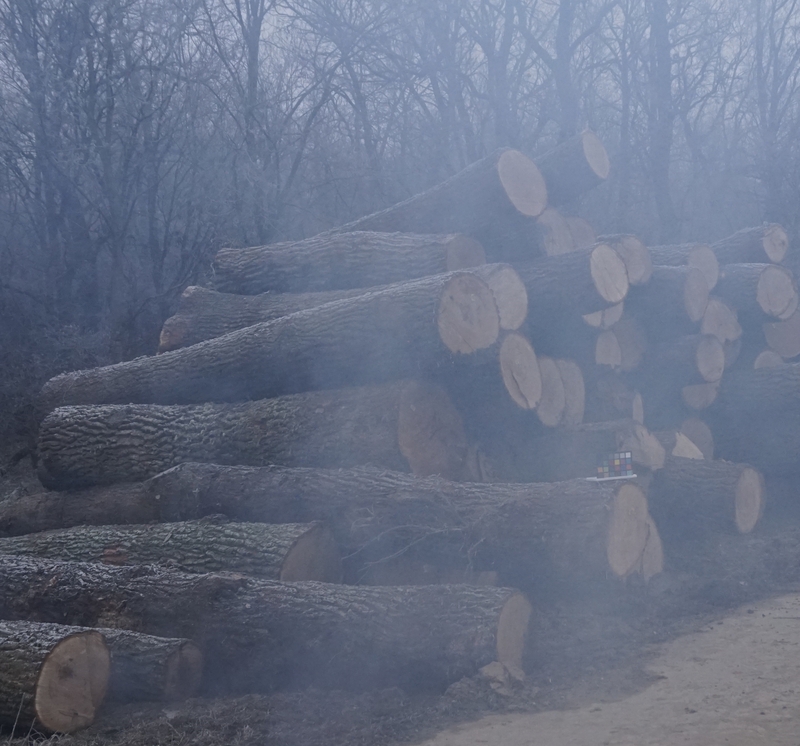}}\hfill
		\subfloat[Uncertainty $\theta$]{\includegraphics[width=0.163\textwidth,height=0.17\textwidth]{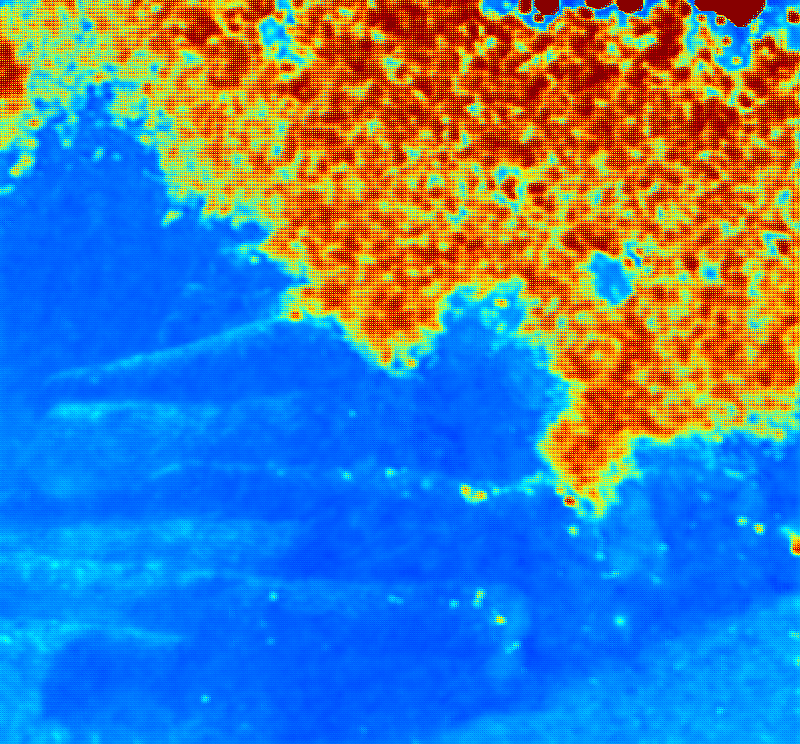}}\hfill
		\subfloat[Output]{\includegraphics[width=0.163\textwidth,height=0.17\textwidth]{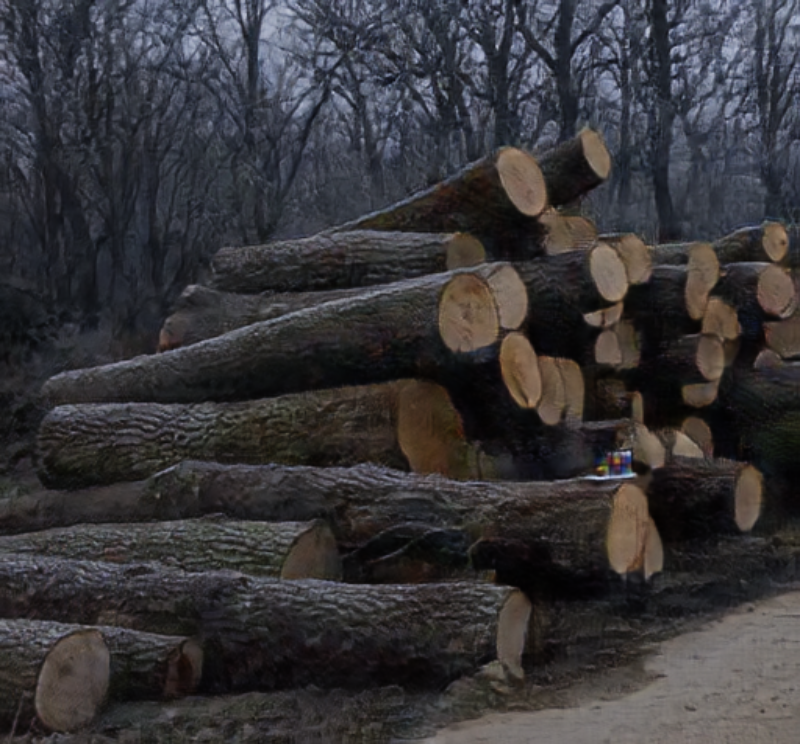}}\hfill
	\caption{Uncertainty maps of O-HAZE~\cite{ancuti2018haze} dataset. The (b) uncertainty map indicates the fog intensity.}
	\label{fig:density}
\end{figure}

\subsection{Uncertainty Feedback Learning}
\begin{figure}[t]
	\centering
	{\includegraphics[width=.9\textwidth]{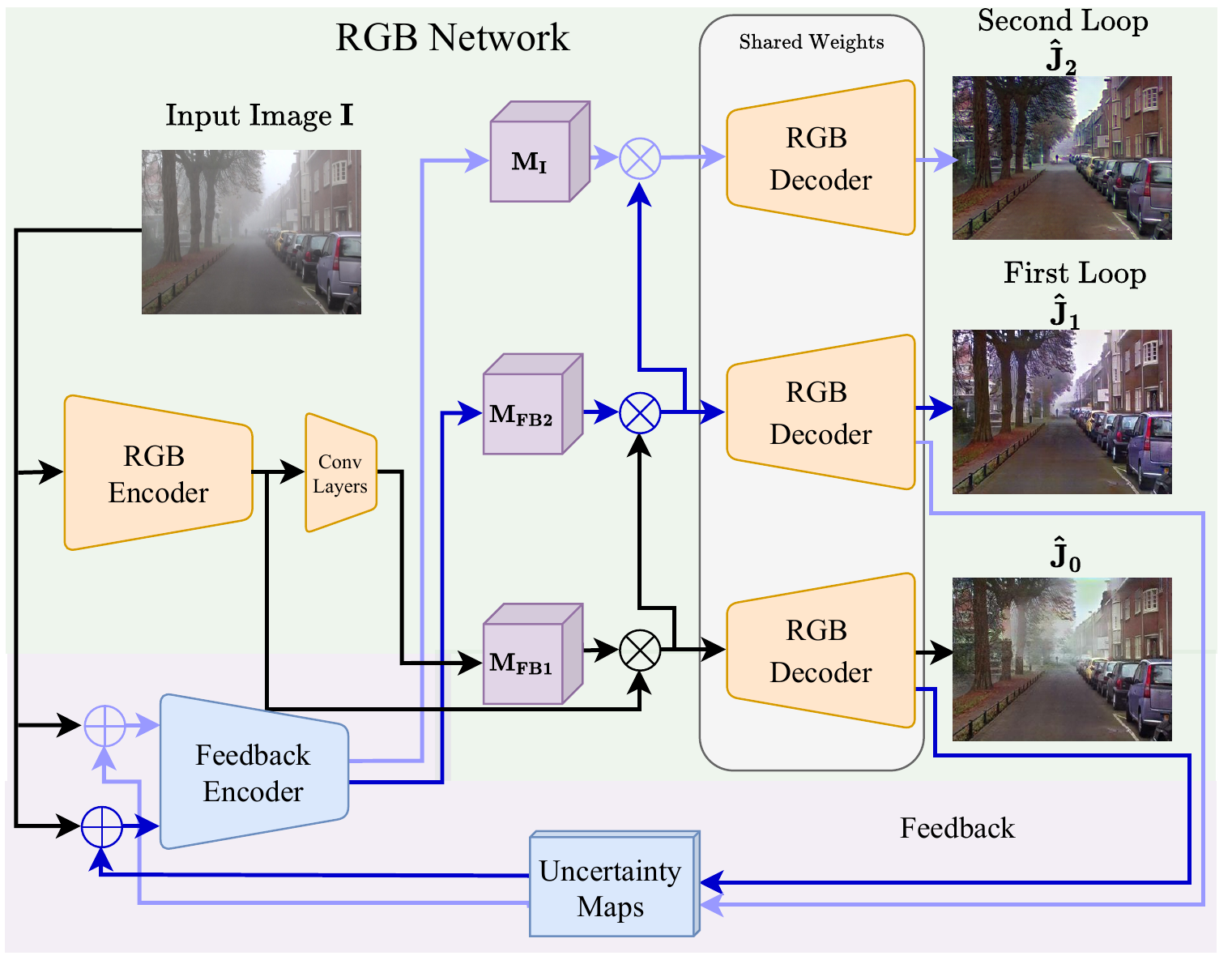}}\hfill
	\caption{Architecture of the uncertainty feedback learning. This network refines the performance of the RGB network.}
	\label{fig:feedback_net}
\end{figure}

\subsubsection{Uncertainty Map}
The main challenge of dealing with dense non-uniform fog distributions is how to differentiate the dense fog regions from the light fog regions.
To address this problem, we exploit an uncertainty map as an attention map to guide the network to differentiate dense fog regions from the rest of the input image. Since the network produces higher uncertainty for the denser fog regions.
Each value in the uncertainty map represents the confidence of the defogging operation at the corresponding pixel (i.e. the variance). The higher the value, the more uncertain the network's prediction for that pixel.

To generate an uncertainty map together with the defogged result, we add a multi-task decoder to our network.
Note that the defogged result and the uncertainty map are decoded from the same features, since there is only one encoder.
We assume that the defogged output $\mathbf{\hat{J}}$ follows a Laplace distribution, where the mean of this distribution is the clear ground truth $\mathbf{J}^{gt}$~\cite{kendall2017uncertainties,ning2021uncertainty}.
Under this assumption, we can define a likelihood function as follows:
\begin{align}
p(\mathbf{J}^{gt} | \mathbf{I}) = \frac{1}{2 \theta }  \exp (-\frac{\left \| \mathbf{\hat{J}} - \mathbf{J}^{gt} \right \|_1}{\theta}),
\label{eq:likelihood}
\end{align}
where $\theta$ is the variance of the Laplace distribution. In our implementation, we define this variance as the uncertainty of the defogged output $\mathbf{\hat{J}}$. Therefore, Eq.~\eqref{eq:likelihood} includes both outputs generated by our multi-task network. Taking the logarithm of both sides of Eq.~\eqref{eq:likelihood} and maximizing it, we can obtain:
$\arg \max_\theta \ln \space p(\mathbf{J}^{gt} | \mathbf{I})  =  -\frac{\left \| \mathbf{\hat{J}} - \mathbf{J}^{gt} \right \|_1}{\theta} -  \ln \theta .$

For the first term in this likelihood $-\frac{\left \| \mathbf{\hat{J}} - \mathbf{J}^{gt} \right \|_1}{\theta}$, we simply convert the negative sign to positive and put it into the loss function. 
The second term  $-\ln \theta$, we convert it to $\ln(\theta + 1)$ to avoid negative infinity when $\theta$ is zero.
Hence, the uncertainty loss we will minimize is expressed as follows:
\begin{align}
\mathcal{L}_{\rm unc} =  \frac{\left \| \mathbf{\hat{J}} - \mathbf{J^{gt}} \right \|_1}{\theta} + \ln (\theta + 1).
\label{eq:uncertainty_synt}
\end{align}

\subsubsection{Uncertainty Feedback Learning}
Unfortunately, the results of our baseline network might still suffer from the remaining fog.
There are two possible reasons.
First, the effectiveness of our multiplier consistency loss depends on the grayscale network's performance. 
While we can see in our ablation studies that this grayscale guidance improves the defogging performance, the grayscale network cannot completely remove fog all the time. 
Second, our discriminative loss cannot fully suppress fog for any input image, since we do not have paired training ground truths for real images.

To address this problem, we introduce uncertainty feedback learning, which the architecture is shown in Fig.~\ref{fig:feedback_net}.
We provide our network extra attention to the different densities of fog based on our network-generated uncertainty maps.
Specifically, we introduce uncertainty feedback learning to make our network focus on areas where fog is still visible and to defog these areas iteratively.
In one iteration, if our output still contains fog in some regions, then the uncertainty map will indicate those regions, and in the next iteration, our method will focus on these regions to further defog them.

As shown in Fig.~\ref{fig:feedback_net}, we feedforward the uncertainty map together with the input image into the feedback encoder, producing a new feedback feature multiplier $\mathbf{M_{FB}}$. 
We multiply this multiplier with the RGB features, and feed the multiplication result to the RGB decoder, generating the enhanced output, $\mathbf{\hat{J}_1}$.
To train our RGB network and the feedback network, we use real images (that do not have ground truths) and apply only the discriminative loss.
We compute the loss of this output $\mathbf{\hat{J}_i}$ (where $i$ is the index of the iterations) with the same loss functions as the initial output image $\mathbf{\hat{J}}$, and backpropagate the errors.
We iterate this process a few times to obtain the final output. 
The number of iterations is constrained by the GPU memory. From our experiments, we found that the uncertainty map tends to be unchanged after two or three iterations.

\subsection{Overall Losses}
In training our network, we use both synthetic images with ground truths and real images without ground truths. 
For both the grayscale and RGB networks, we feedforward a set of synthetic images into the network, which outputs the predicted clear synthetic images.
For the same batch, we feedforward a set of real images into the network, producing the predicted real images.
Having obtained the predicted clear images of both the synthetic and real images, we then train the discriminators in the grayscale channel of the grayscale and RGB networks.
Training discriminators requires a set of reference images, which must be real and clear (without fog).
Our reference images include the ground truth images of the synthetic fog images (paired), and other real images that with no correlation to our input images (unpaired).

We multiply each loss function with its respective weight, and sum them together to obtain our overall loss function: 
\begin{equation}
\mathcal{L} = \lambda_{\rm m} \mathcal{L}_{\rm multiplier} + \lambda_{\rm s} \mathcal{L}_{\rm structure} + \lambda_{\rm u} \mathcal{L}_{\rm unc} +
\mathcal{L}_{\rm MSE} + \lambda_{\rm d} \mathcal{L}_{\rm dis}, 
\label{eq:losses_more}
\end{equation}
where $\lambda$ are the weights for the respective losses. 
$\lambda_{\rm m}$ = 1, $\lambda_{\rm u}$ = 1, their values are obtained empirically.
$\lambda_{\rm s}$ = 0.1, $\lambda_{\rm d}$ = 0.005, their values are followed default setting. 
$\mathcal{L}_{\rm MSE}$ is the Mean Squared Error (MSE) loss (applied only to synthetic images), 
$\mathcal{L}_{\rm dis}$ is the discriminative loss.

\section{Experimental Results}
\label{sec:exp}

\begin{figure*}[t!]
	\centering
	\captionsetup[subfloat]{farskip=1pt}
	\captionsetup[subfigure]{labelformat=empty}
	\setcounter{subfigure}{0}
	\subfloat[Input]{\includegraphics[width = 0.196\textwidth]{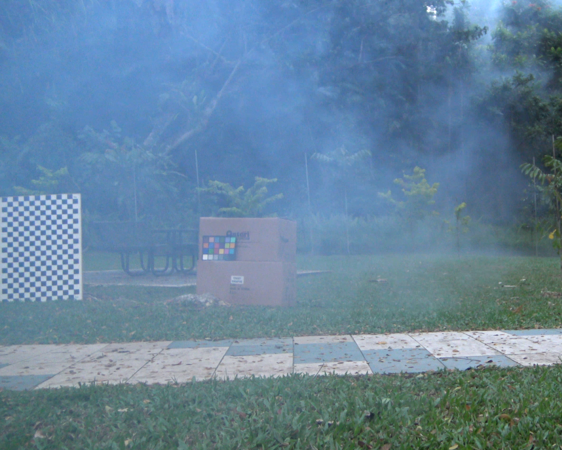}}\hfill
	\subfloat[Ours]{\includegraphics[width = 0.196\textwidth]{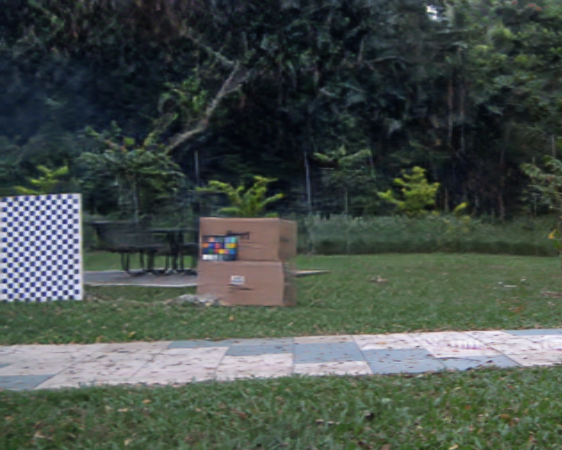}}\hfill
	\subfloat[Dehamer'22~\cite{guo2022image}]{\includegraphics[width = 0.196\textwidth]{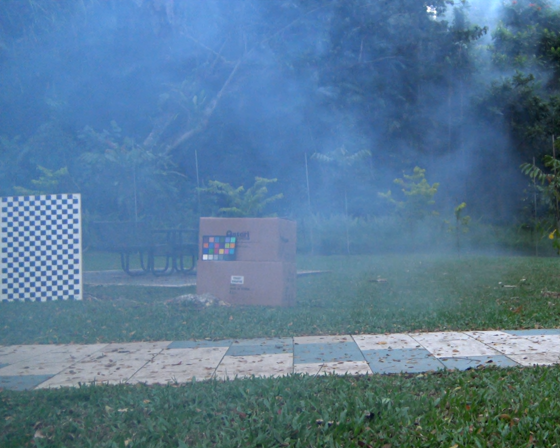}}\hfill
	\subfloat[DehazeF.'22~\cite{song2022vision}]{\includegraphics[width = 0.196\textwidth]{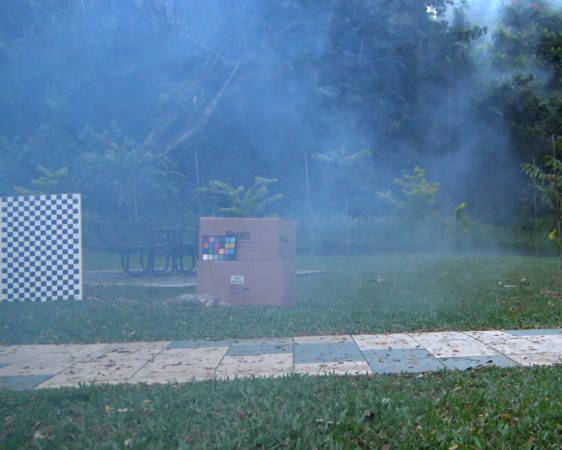}}\hfill
	\subfloat[D4'22~\cite{yang2022self}]{\includegraphics[width = 0.196\textwidth]{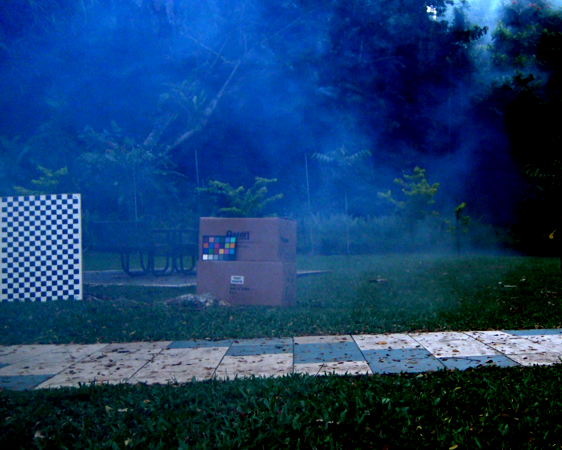}}\hfill
	\subfloat[PSD'21~\cite{chen2021psd}]{\includegraphics[width = 0.196\textwidth]{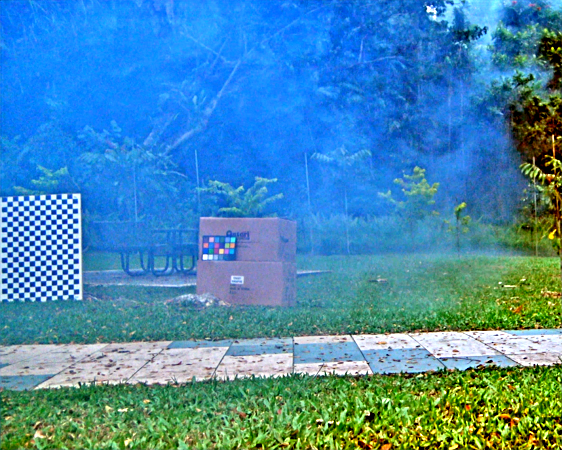}}\hfill
	\subfloat[4K'21~\cite{zheng2021ultra}]{\includegraphics[width = 0.196\textwidth]{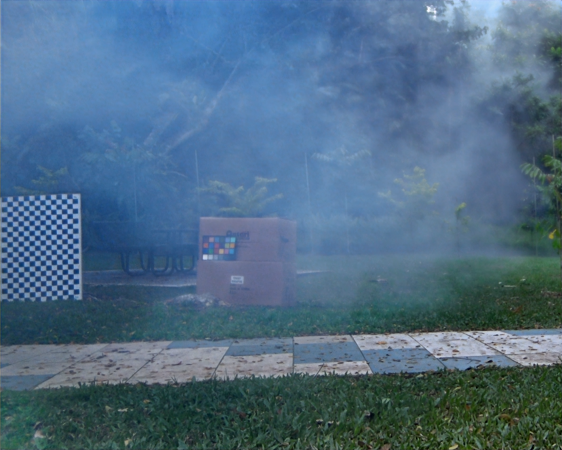}}\hfill
	\subfloat[MSBDN~\cite{dong2020multi}]{\includegraphics[width = 0.196\textwidth]{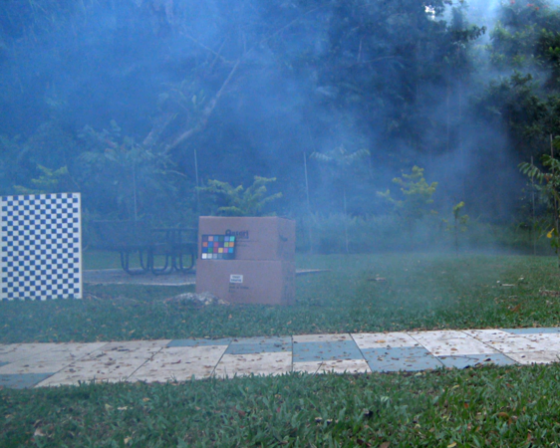}}\hfill
	\subfloat[DAN~\cite{shao2020domain}]{\includegraphics[width = 0.196\textwidth]{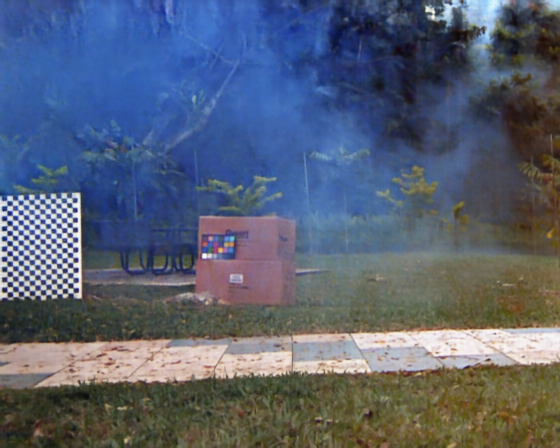}}\hfill
	\subfloat[GDN~\cite{liu2019griddehazenet}]{\includegraphics[width = 0.196\textwidth]{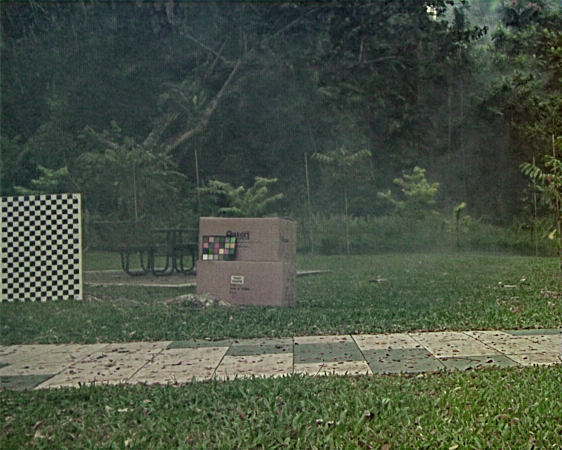}}\hfill
	\setcounter{subfigure}{0}
	\subfloat[Input]{\includegraphics[width = 0.196\textwidth]{fig/trailer/in.png}}\hfill
	\subfloat[Ours]{\includegraphics[width = 0.196\textwidth]{fig/trailer/our.png}}\hfill
	\subfloat[Dehamer'22~\cite{guo2022image}]{\includegraphics[width = 0.196\textwidth]{fig/trailer/DeHamer.png}}\hfill
	\subfloat[DehazeF.'22~\cite{song2022vision}]{\includegraphics[width = 0.196\textwidth]{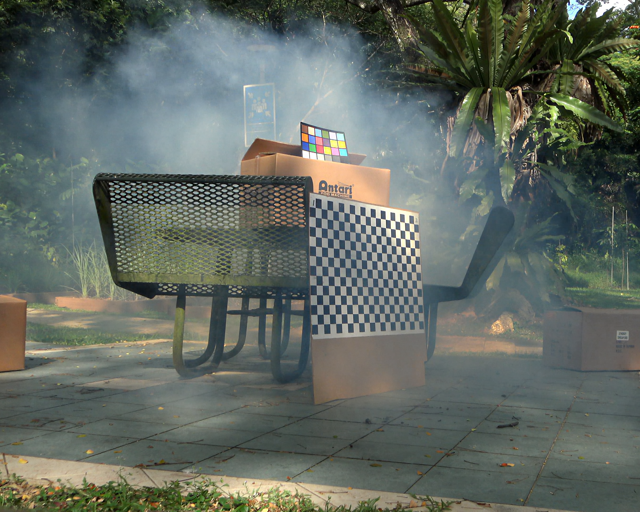}}\hfill
	\subfloat[D4'22~\cite{yang2022self}]{\includegraphics[width = 0.196\textwidth]{fig/trailer/D4.png}}\hfill
	\subfloat[PSD'21~\cite{chen2021psd}]{\includegraphics[width = 0.196\textwidth]{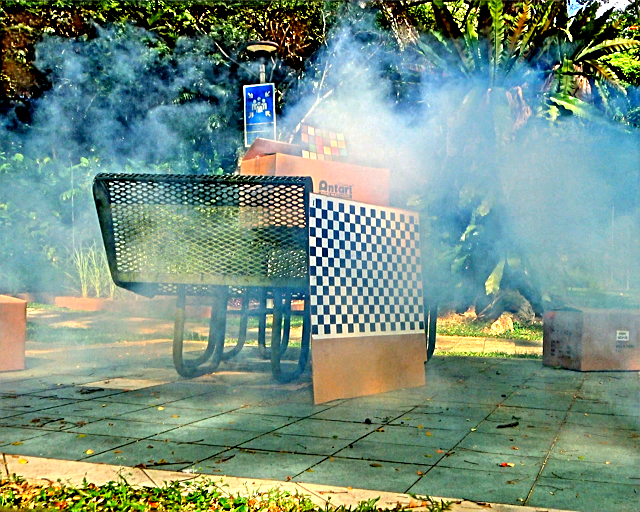}}\hfill
	\subfloat[4K'21~\cite{zheng2021ultra}]{\includegraphics[width = 0.196\textwidth]{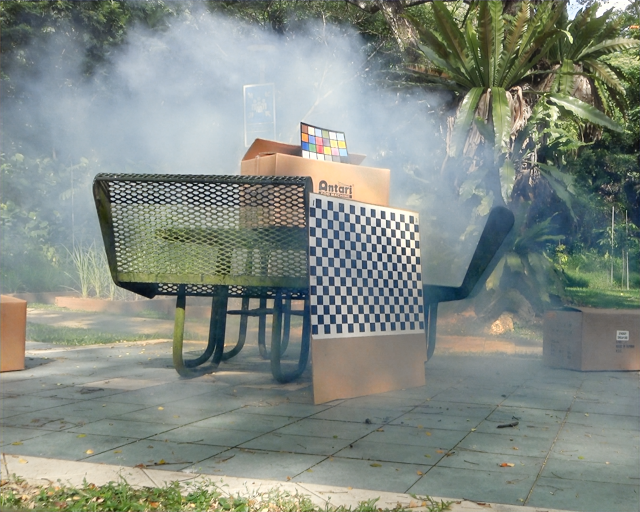}}\hfill
	\subfloat[MSBDN~\cite{dong2020multi}]{\includegraphics[width = 0.196\textwidth]{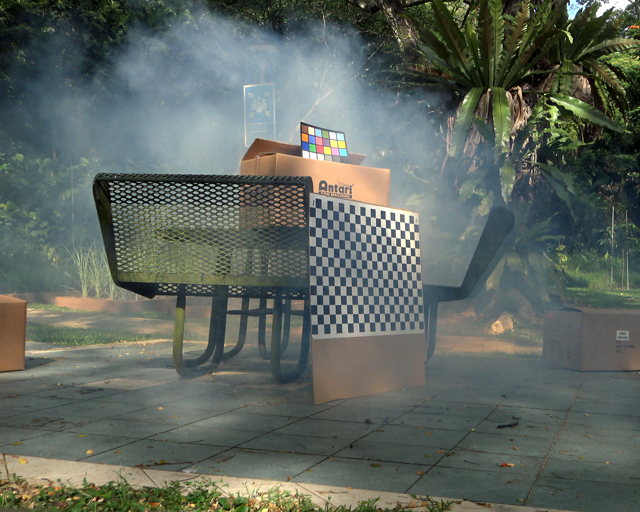}}\hfill
	\subfloat[DAN~\cite{shao2020domain}]{\includegraphics[width = 0.196\textwidth]{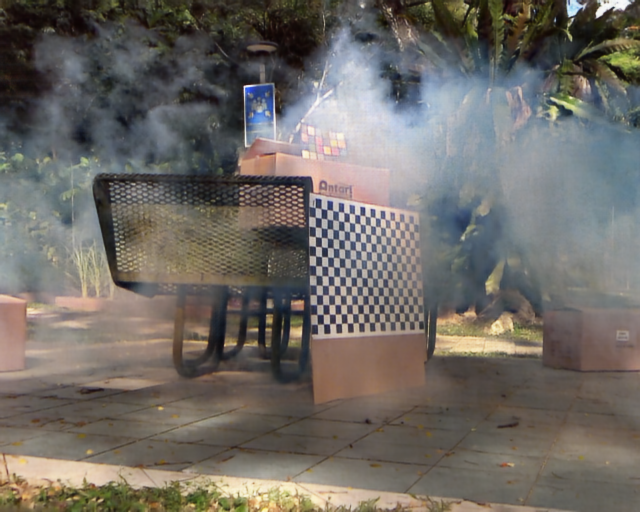}}\hfill
	\subfloat[GDN~\cite{liu2019griddehazenet}]{\includegraphics[width = 0.196\textwidth]{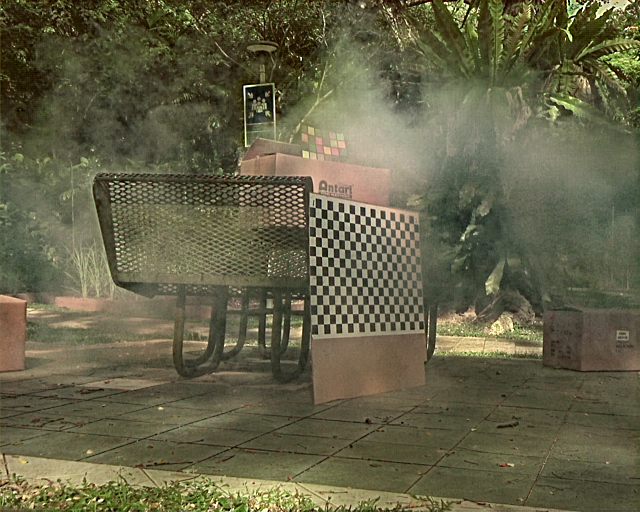}}\hfill
	\setcounter{subfigure}{0}
	\subfloat[Input]{\includegraphics[width=0.196\textwidth]{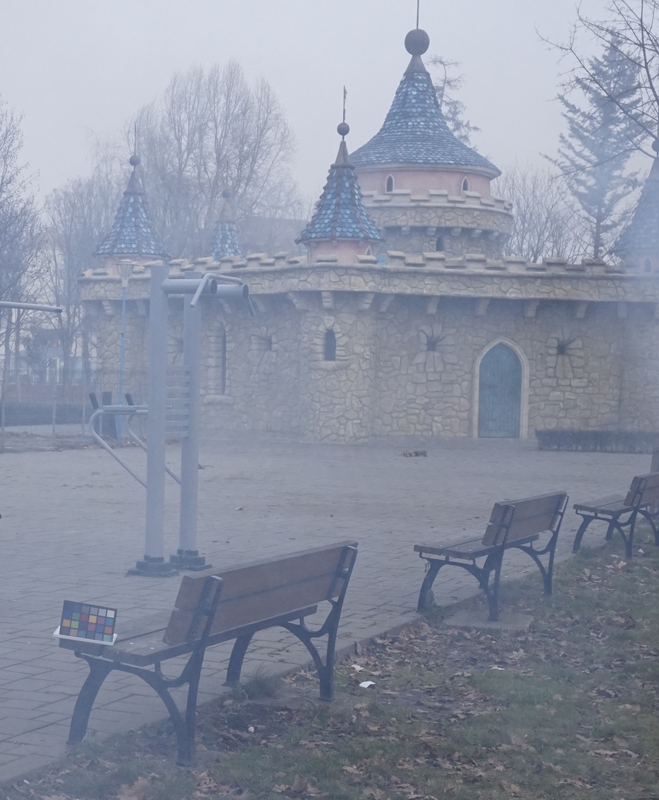}}\hfill
	\subfloat[Ours]{\includegraphics[width=0.196\textwidth]{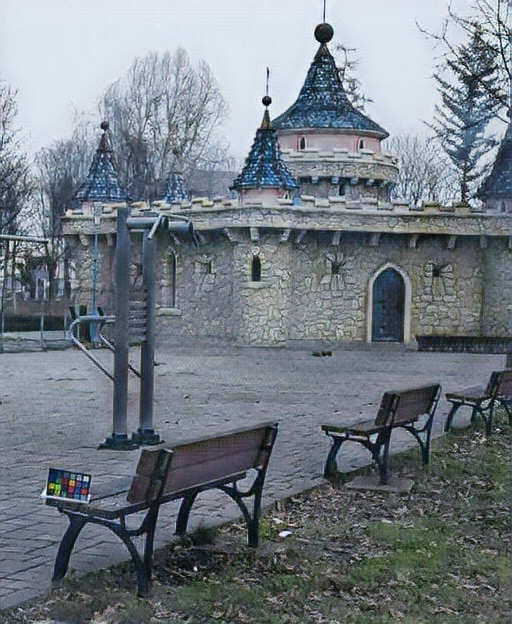}}\hfill
	\subfloat[Dehamer'22~\cite{guo2022image}]{\includegraphics[width = 0.196\textwidth]{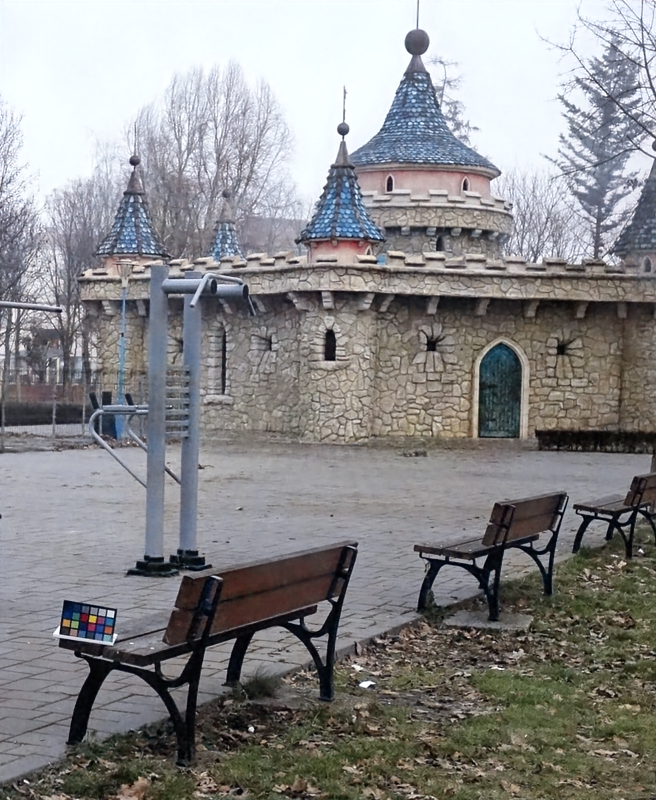}}\hfill
	\subfloat[DehazeF.'22~\cite{song2022vision}]{\includegraphics[width = 0.196\textwidth]{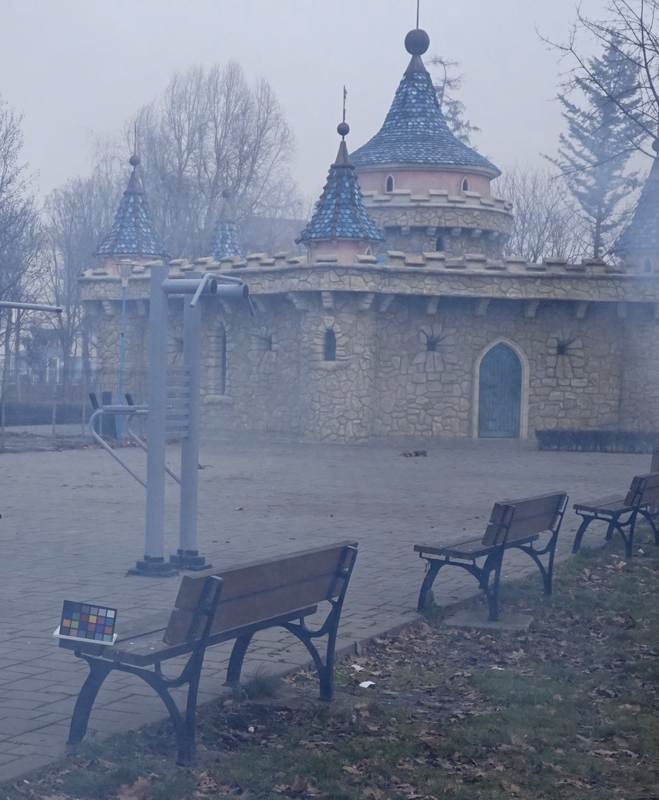}}\hfill
	\subfloat[D4'22~\cite{yang2022self}]{\includegraphics[width = 0.196\textwidth]{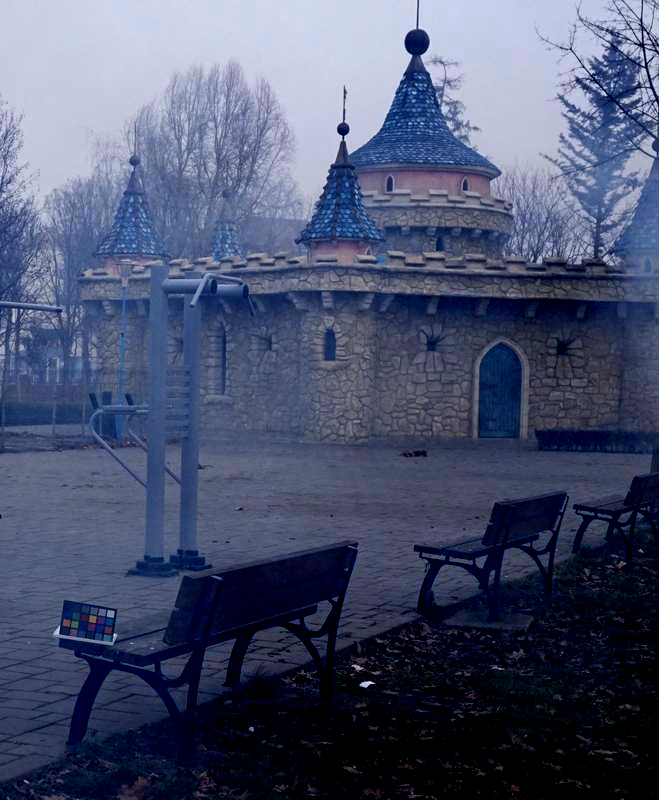}}\hfill
	\subfloat[PSD'21~\cite{chen2021psd}]{\includegraphics[width=0.196\textwidth]{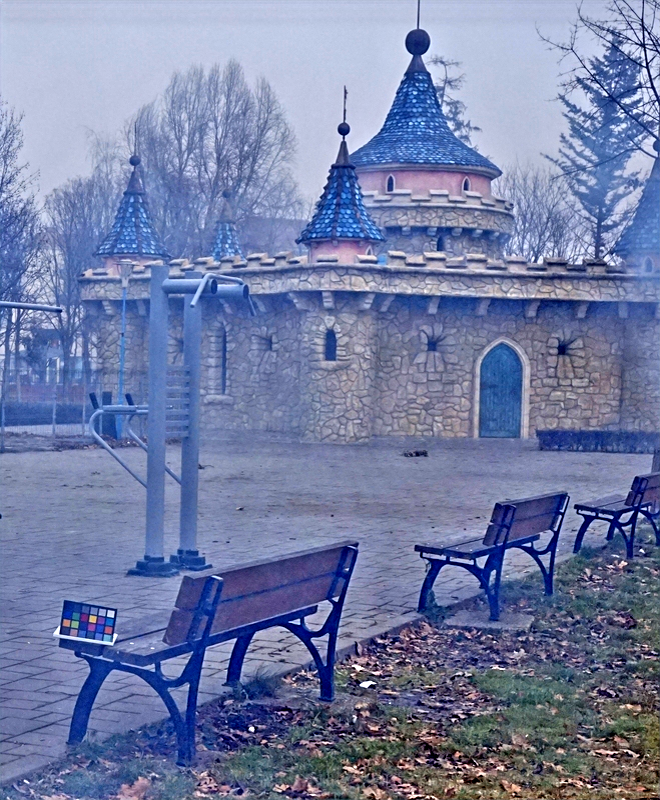}}\hfill
	\subfloat[4K'21~\cite{zheng2021ultra}]{\includegraphics[width=0.196\textwidth]{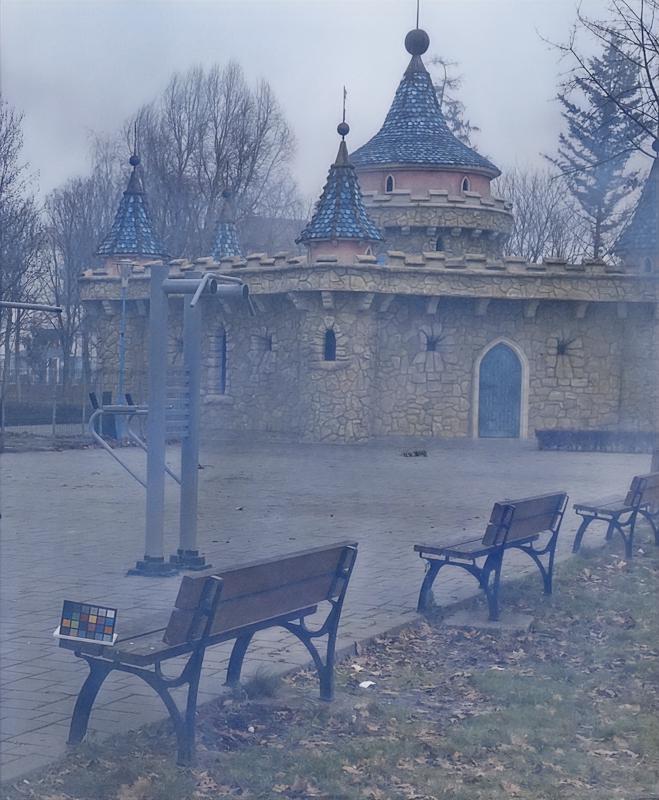}}\hfill
	\subfloat[MSBDN~\cite{dong2020multi}]{\includegraphics[width=0.196\textwidth]{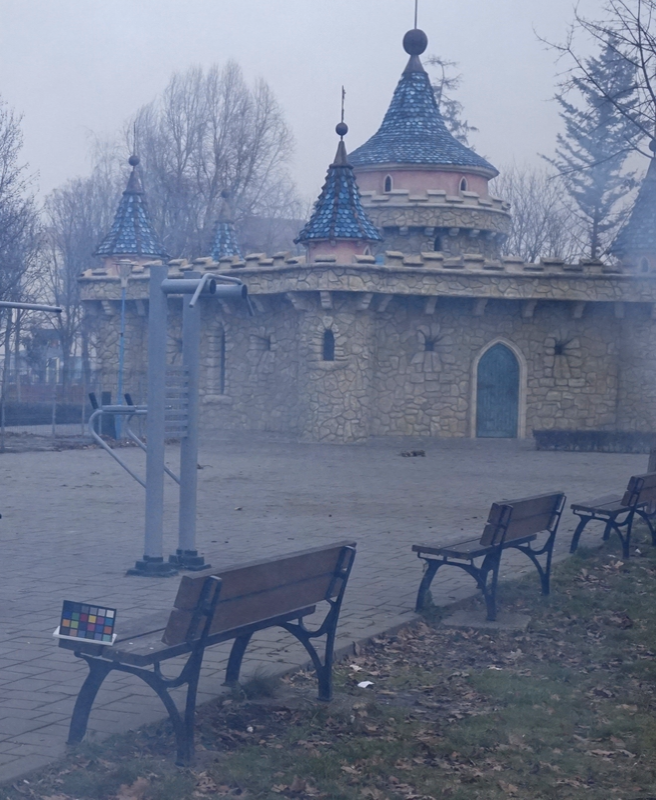}}\hfill
	\subfloat[DAN~\cite{shao2020domain}]{\includegraphics[width=0.196\textwidth]{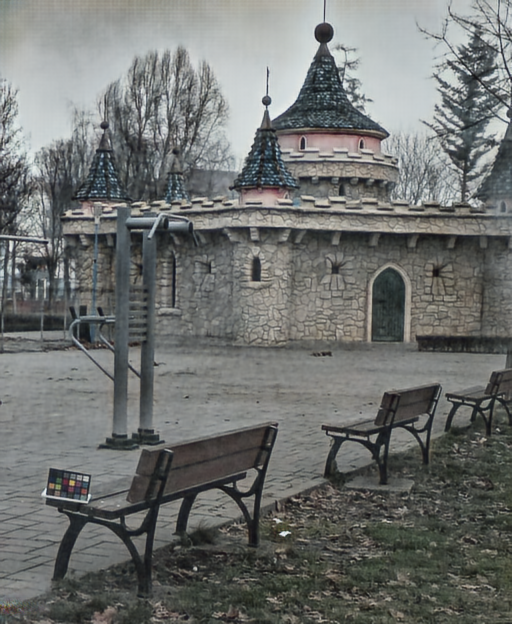}}\hfill
	\subfloat[GDN~\cite{liu2019griddehazenet}]{\includegraphics[width = 0.196\textwidth]{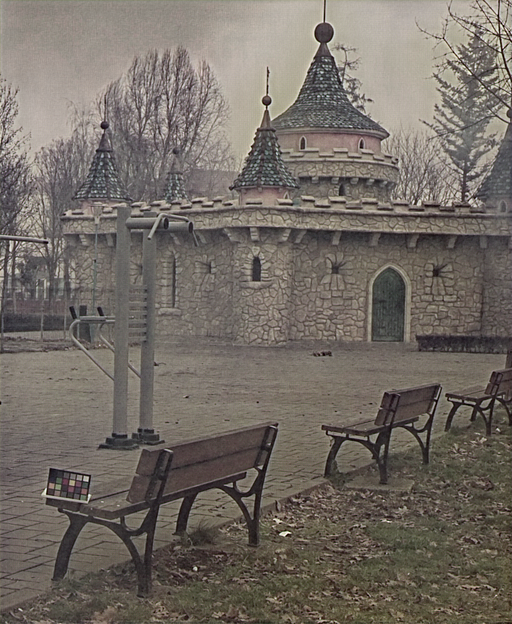}}\hfill
	\caption{Qualitative evaluation results on real fog machine and O-Haze~\cite{ancuti2018haze} images.}
	\label{fig:ohaze}
\end{figure*}

\noindent \textbf{Implementation}
We use two sets of data to train our networks: real fog images and reference clear images, synthetic fog images and their ground truth.
For the real fog images, we train on self-collected and Internet fog images.
For the clear reference images, we collect clear images from Google Street View and Internet.
For synthetic training images, we render fog images (Eq.~\ref{eq:asmodel}) from clear images, taken from Cityscapes~\cite{cordts2016cityscapes}, which provides 2,975 pairs of RGB images and their disparity maps.
Subsequently, we fine-tune the model on different datasets. 
For self-collected smoke images, we fine-tune the model on the 110 self-collected smoke images and clean pairs, and 100 unpaired Internet clean references.
We also collect 12 other pairs of fog data for evaluation. Our data is publicly available.

\begin{table}[t!]
	\centering
	\caption{Quantitative results on Dense-HAZE, NH-HAZE, O-HAZE and self-collected smoke datasets.}
	\resizebox{0.9\columnwidth}{!}{
	\begin{tabular}{l|c|c|c|c|c|c|c|c}
		\toprule[1.2 pt]
		\multirow{2}{*}{Method}
		&       \multicolumn{2}{c|}{Dense-HAZE \cite{ancuti2019dense}}
		&       \multicolumn{2}{c|}{NH-HAZE \cite{ancuti2020nh}}
		&       \multicolumn{2}{c|}{O-HAZE \cite{ancuti2018haze}} 
		&       \multicolumn{2}{c}{SMOKE} 
		\\[1pt]
		& {PSNR$\uparrow$}  & {SSIM$\uparrow$}  &{PSNR$\uparrow$}  & {SSIM$\uparrow$} &{PSNR$\uparrow$}  & {SSIM$\uparrow$} &{PSNR$\uparrow$}  & {SSIM$\uparrow$}
		\\[1pt] \hline
		DCP \cite{he2010single}           &10.06 &0.39 &10.57 &0.52 &16.78 &0.65 &11.26 &0.26\\[1pt]
		DehazeNet \cite{cai2016dehazenet} &13.84 &0.43 &16.62 &0.52 &17.57 &0.77 &- &-\\[1pt]
		AODNet \cite{li2017aod}           &13.14 &0.41 &15.40 &0.57 &15.03 &0.54 &- &-\\[1pt]
		GDN \cite{liu2019griddehazenet}   &-     &-    &-     &-    &23.51 &\bf{0.83} &15.19 &0.53\\[1pt]
		MSBDN \cite{dong2020multi}        &15.37 &0.49 &19.23 &0.71 &24.36 &0.75 &13.19	 &0.34     \\[1pt]
		FFA-Net \cite{qin2020ffa}         &14.39 &0.45 &19.87 &0.69 &22.12 &0.77 &- &-   \\[1pt]
		AECR-Net \cite{wu2021contrastive} &15.80 &0.47 &19.88 &\bf{0.72} &- &-   &- &-   \\[1pt]\hline
		DeHamer'22~\cite{guo2022image}    &16.62 &\bf{0.56} &20.66 &0.68 &17.02 &0.43 &13.31 &0.28     \\[1pt]\hline
		\textbf{Ours}                     &\bf{16.67} &0.50 &\bf{20.99} &0.61 &\bf{24.61} &0.75 &\bf{18.83} &\bf{0.62}\\[1pt] 
		\bottomrule[1.2 pt]
	\end{tabular}
	}
	\label{tab:comparisons}
\end{table}

\subsubsection{Datasets}
We collected real smoke data by ourselves. We used a fog machine to generate fog, where we fixed the camera pose to record fog images and their paired ground truth.
Ancuti et al.~\cite{ancuti2018haze} propose the O-HAZE dataset consisting of 45 pairs of hazy/clear scenes using a smoke generator to simulate the atmospheric scattering effect in the real world. 
Using the same smoke generator equipment, Ancuti et al.~\cite{ancuti2019dense} also propose the Dense-HAZE and NH-HAZE~\cite{ancuti2020nh,ancuti2020ntire}, which both consist of 55 pairs of hazy/clear scenes, 45 training, 5 validation and 5 test images.
The scenes in the Dense-HAZE and NH-HAZE datasets are similar to the O-Haze dataset, but the smoke density is much higher and more non-homogeneous.

\begin{figure*}[t]
	\centering
	\captionsetup[subfigure]{font=small, labelformat=empty}
	\captionsetup[subfloat]{farskip=2pt}
	\centering
	\setcounter{subfigure}{0}
	\subfloat[Input]{\includegraphics[width = 0.196\textwidth]{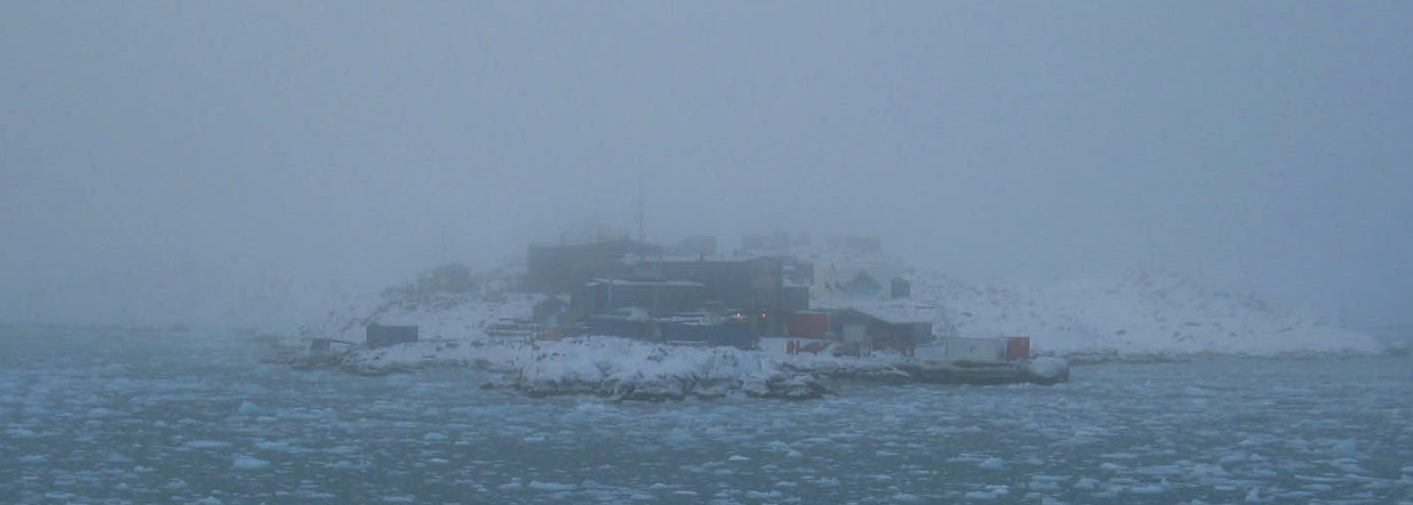}}\hfill
	\subfloat[Ours]{\includegraphics[width = 0.196\textwidth]{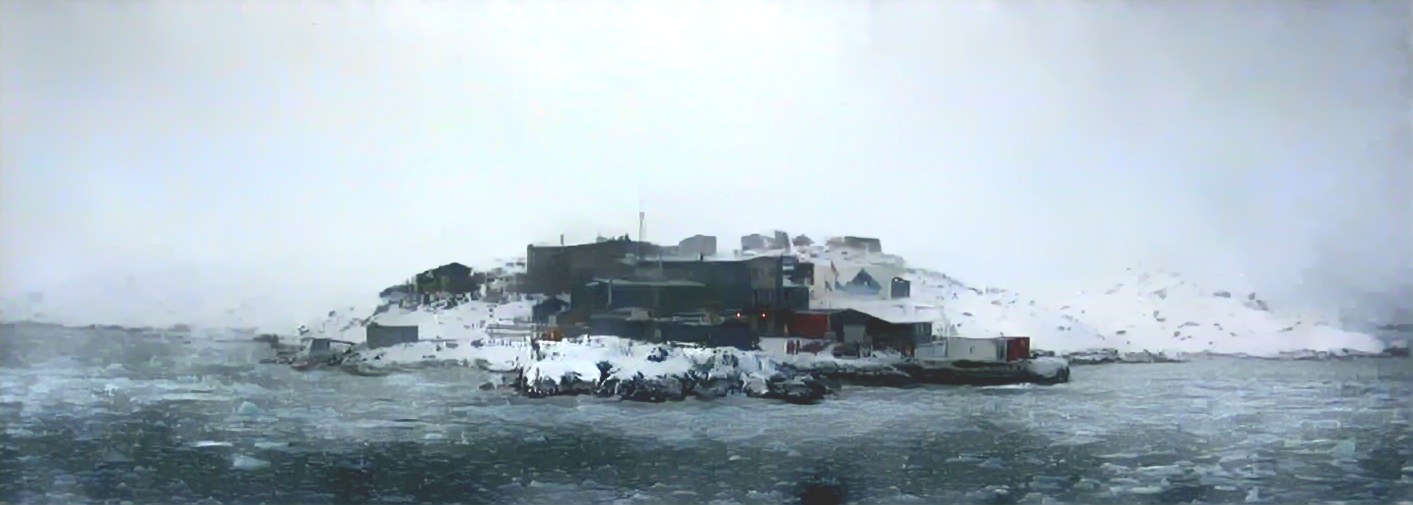}}\hfill
	\subfloat[Dehamer'22~\cite{guo2022image}]{\includegraphics[width = 0.196\textwidth]{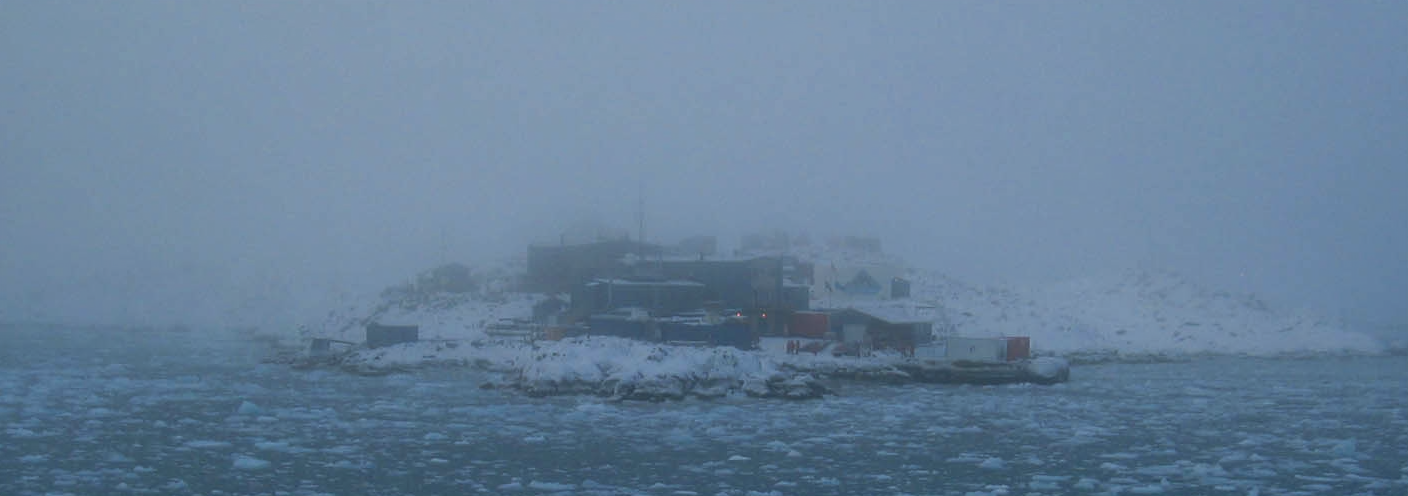}}\hfill
	\subfloat[DehazeF.'22~\cite{song2022vision}]{\includegraphics[width = 0.196\textwidth]{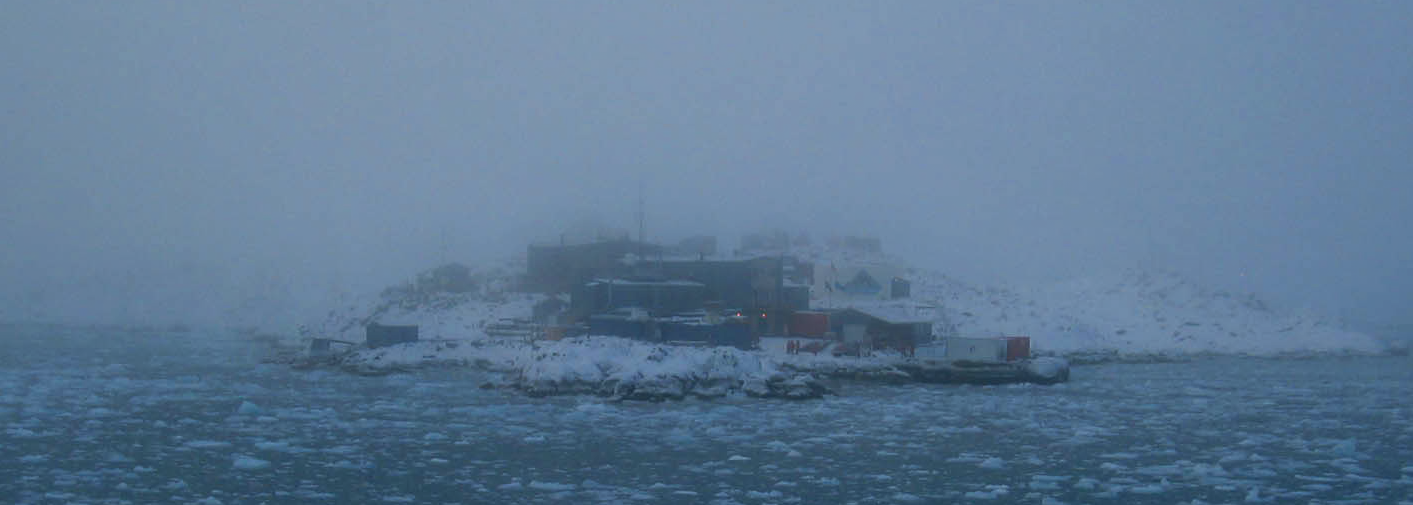}}\hfill
	\subfloat[D4'22~\cite{yang2022self}]{\includegraphics[width = 0.196\textwidth]{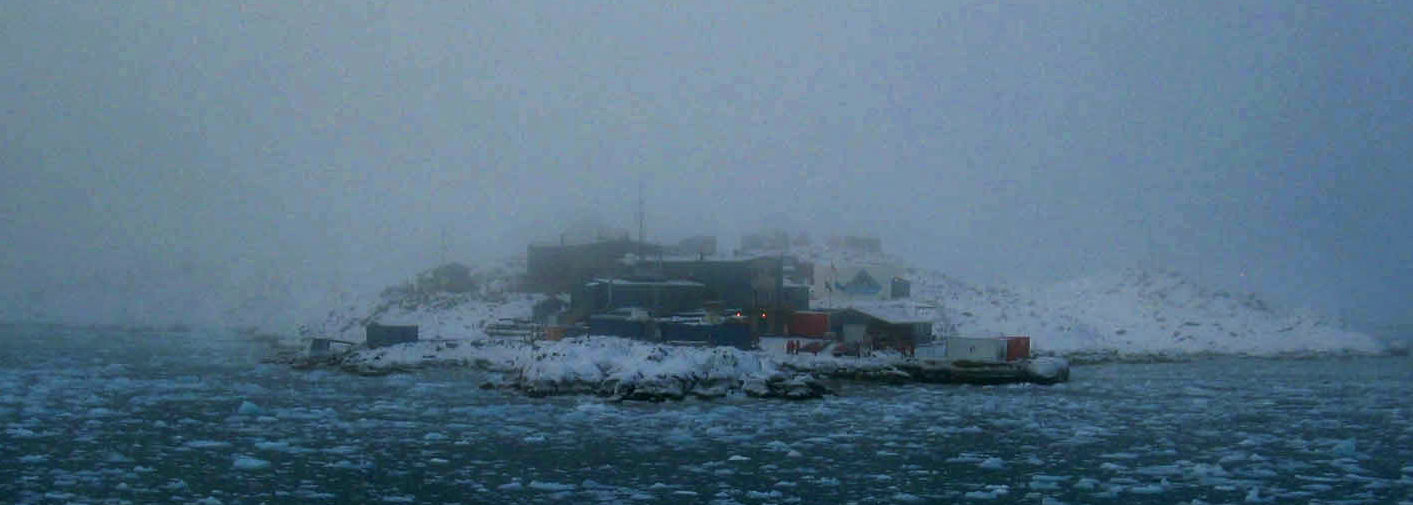}}\hfill
	\subfloat[PSD'21~\cite{chen2021psd}]{\includegraphics[width = 0.196\textwidth]{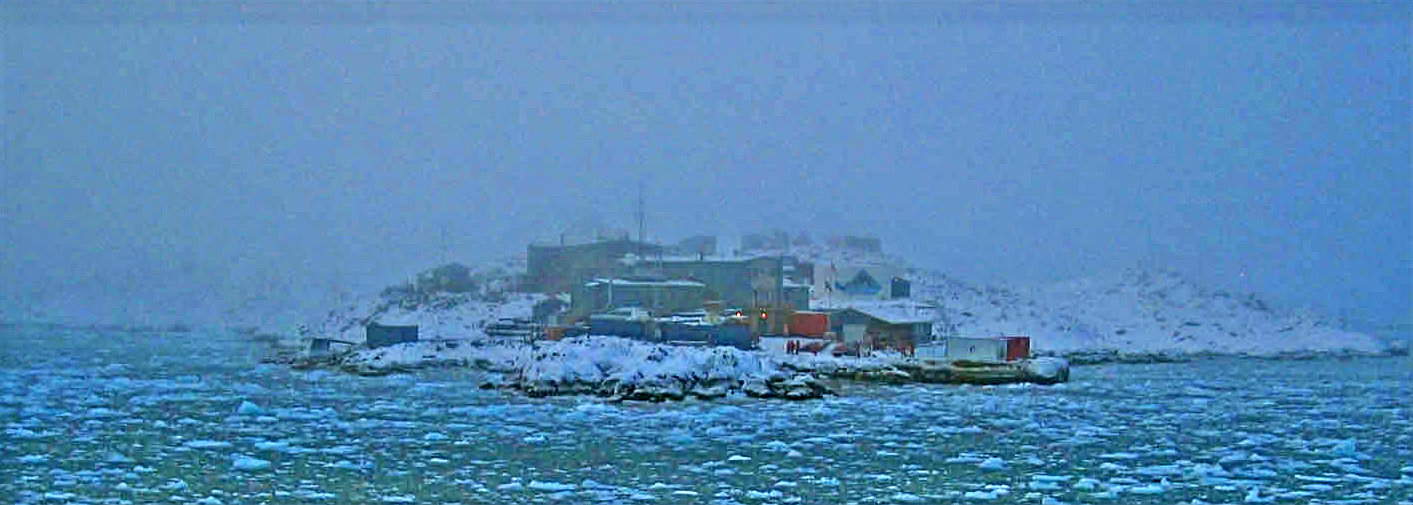}}\hfill
	\subfloat[4K'21~\cite{zheng2021ultra}]{\includegraphics[width = 0.196\textwidth]{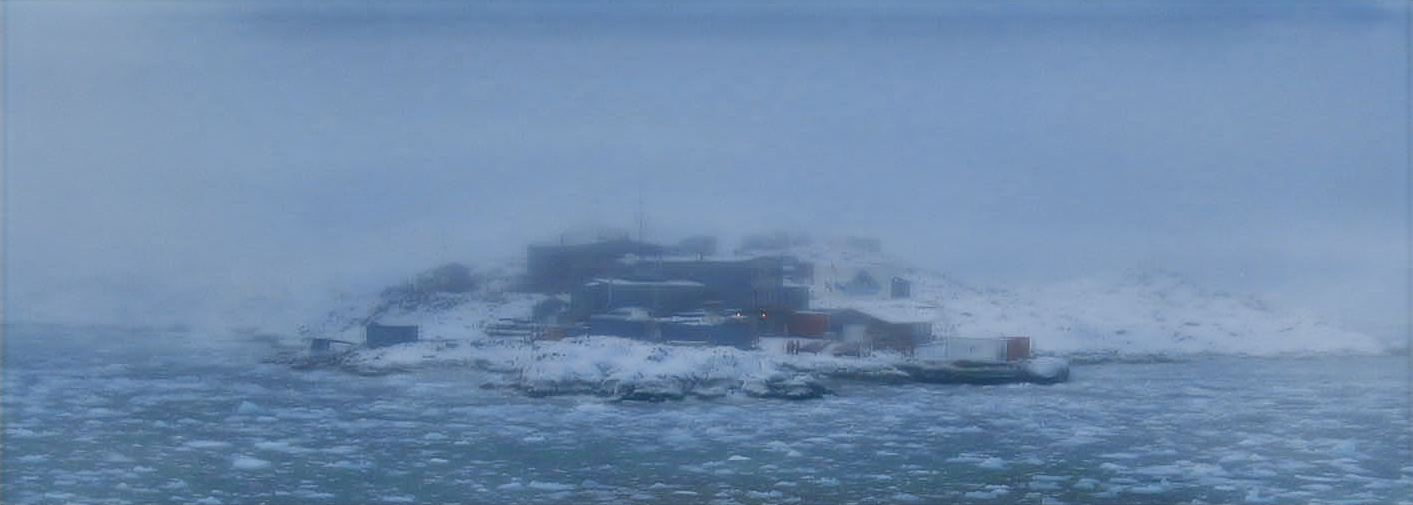}}\hfill
	\subfloat[MSBDN~\cite{dong2020multi}]{\includegraphics[width = 0.196\textwidth]{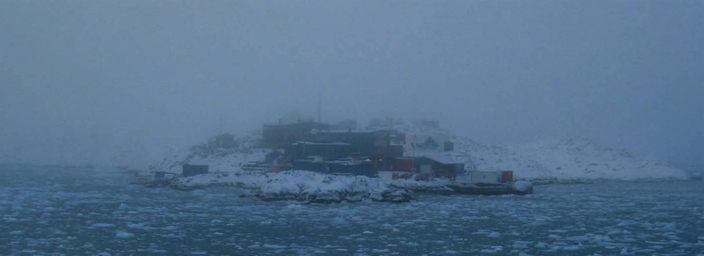}}\hfill
	\subfloat[DAN~\cite{shao2020domain}]{\includegraphics[width = 0.196\textwidth]{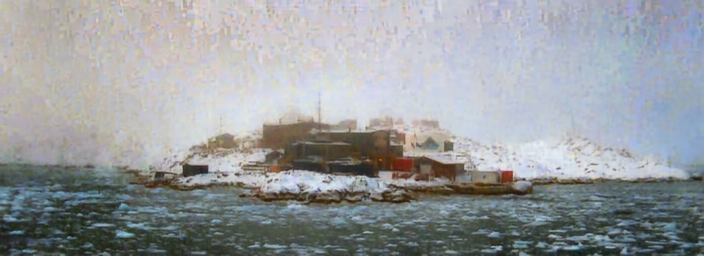}}\hfill
	\subfloat[GDN~\cite{liu2019griddehazenet}]{\includegraphics[width = 0.196\textwidth]{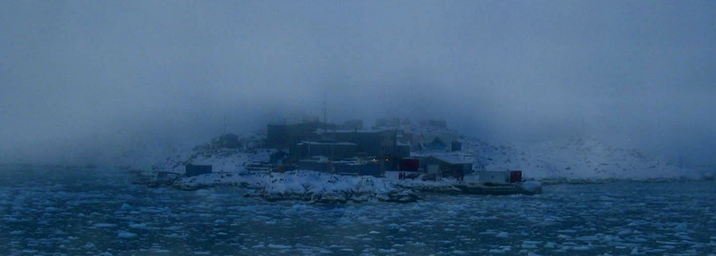}}\hfill	
	\setcounter{subfigure}{0}
	\subfloat[Input]{\includegraphics[width = 0.196\textwidth]{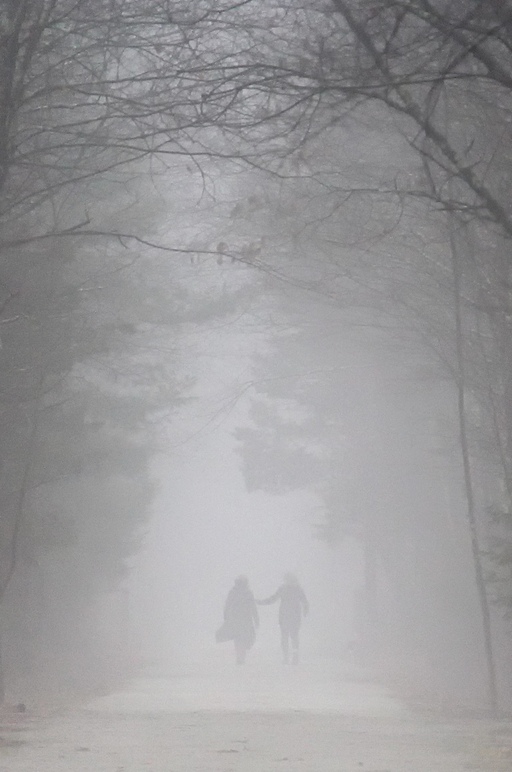}}\hfill
	\subfloat[Ours]{\includegraphics[width = 0.196\textwidth]{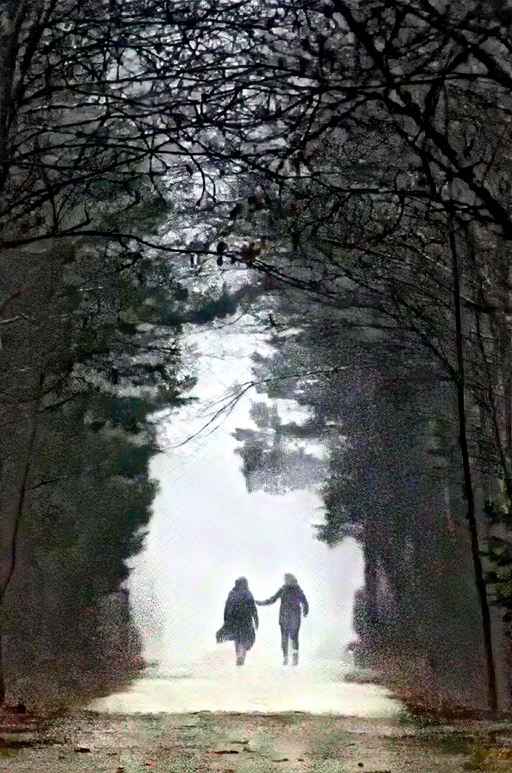}}\hfill
	\subfloat[Dehamer'22~\cite{guo2022image}]{\includegraphics[width = 0.196\textwidth]{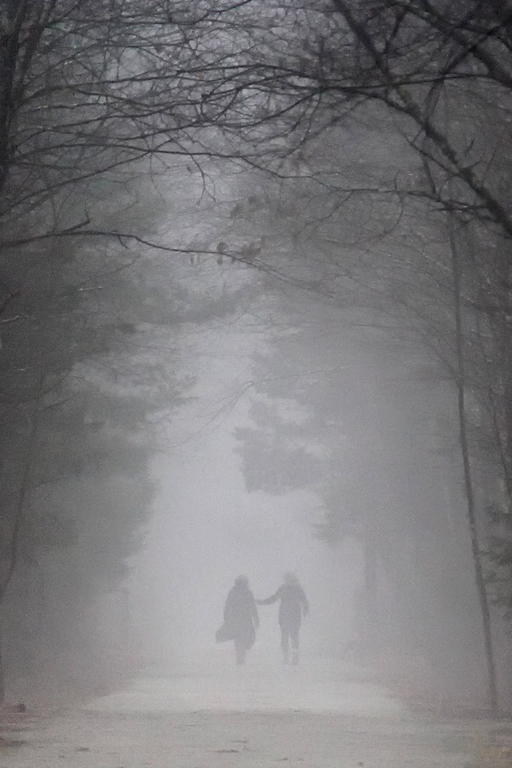}}\hfill
	\subfloat[DehazeF.'22~\cite{song2022vision}]{\includegraphics[width = 0.196\textwidth]{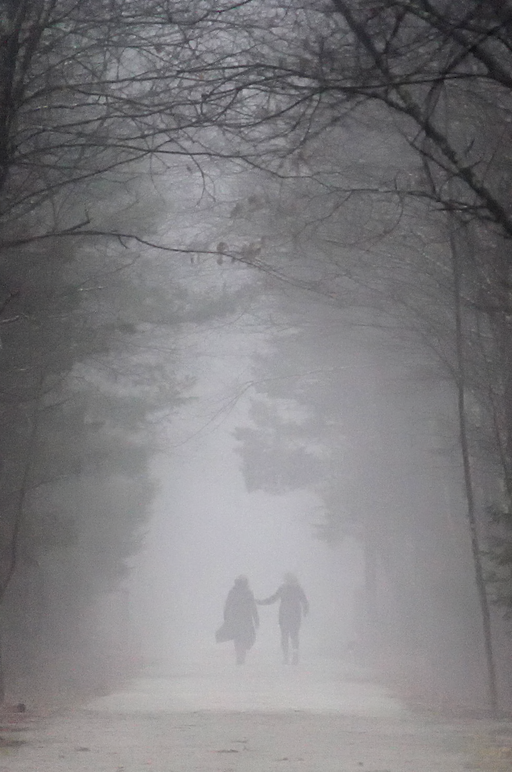}}\hfill
	\subfloat[D4'22~\cite{yang2022self}]{\includegraphics[width = 0.196\textwidth]{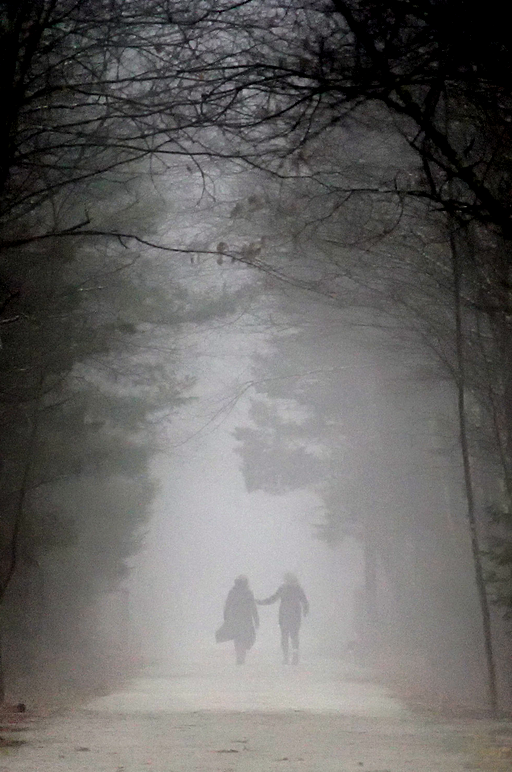}}\hfill
	\subfloat[PSD'21~\cite{chen2021psd}]{\includegraphics[width = 0.196\textwidth]{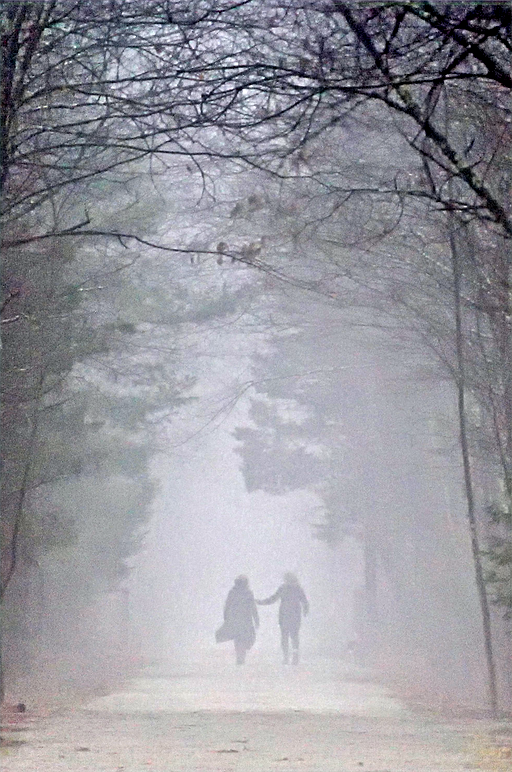}}\hfill
	\subfloat[4K'21~\cite{zheng2021ultra}]{\includegraphics[width = 0.196\textwidth]{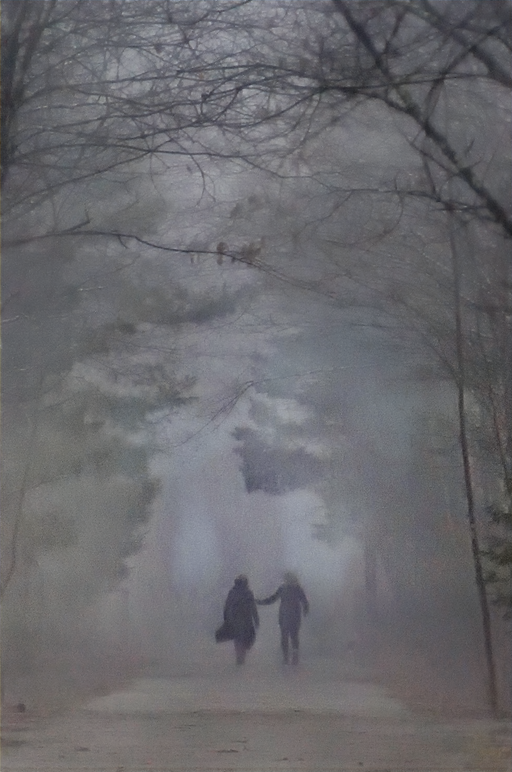}}\hfill
	\subfloat[MSBDN~\cite{dong2020multi}]{\includegraphics[width = 0.196\textwidth]{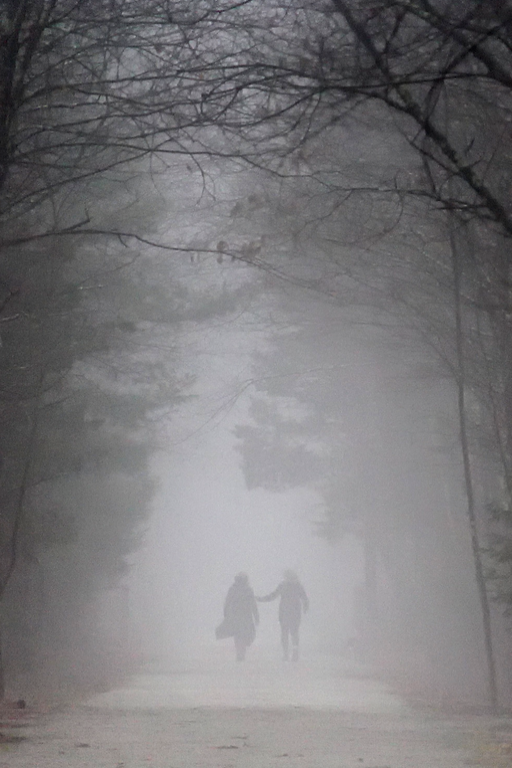}}\hfill
	\subfloat[DAN~\cite{shao2020domain}]{\includegraphics[width = 0.196\textwidth]{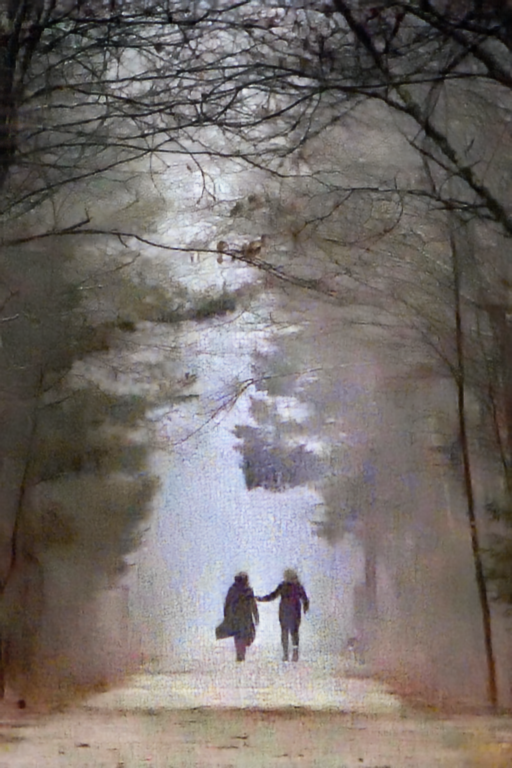}}\hfill
	\subfloat[GDN~\cite{liu2019griddehazenet}]{\includegraphics[width = 0.196\textwidth]{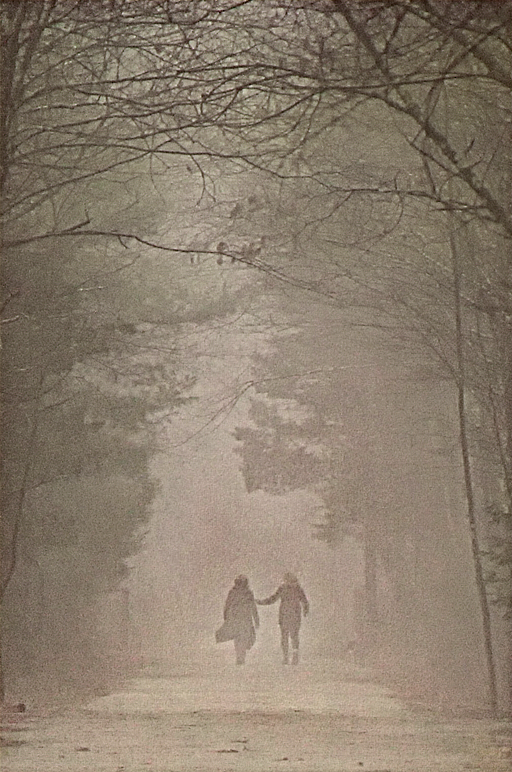}}\hfill
	\caption{Comparison results on commonly used test foggy images.
		(a) Input images. (b) Our results. (c)$\sim$(g) Results of the state-of-the-art methods.}
	\label{fig:real1}
\end{figure*}

\begin{figure*}[t]
	\centering
	\captionsetup[subfigure]{font=small, labelformat=empty}
	\captionsetup[subfloat]{farskip=2pt}
	\centering
	\setcounter{subfigure}{0}
	\subfloat[Input]{\includegraphics[width = 0.196\textwidth]{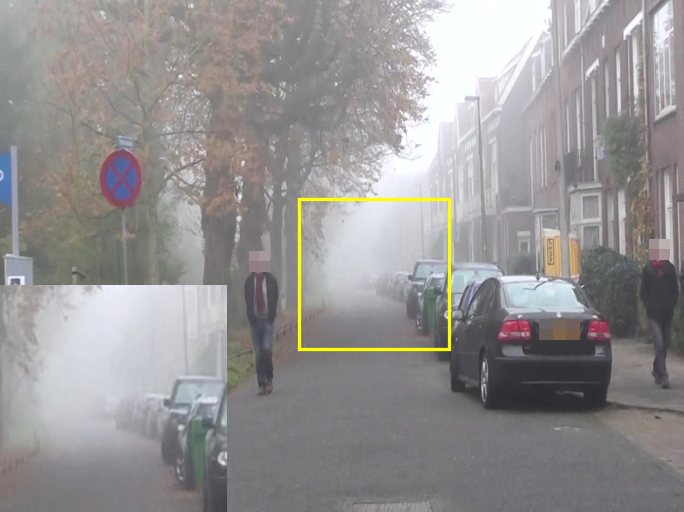}}\hfill
	\subfloat[Ours]{\includegraphics[width = 0.196\textwidth]{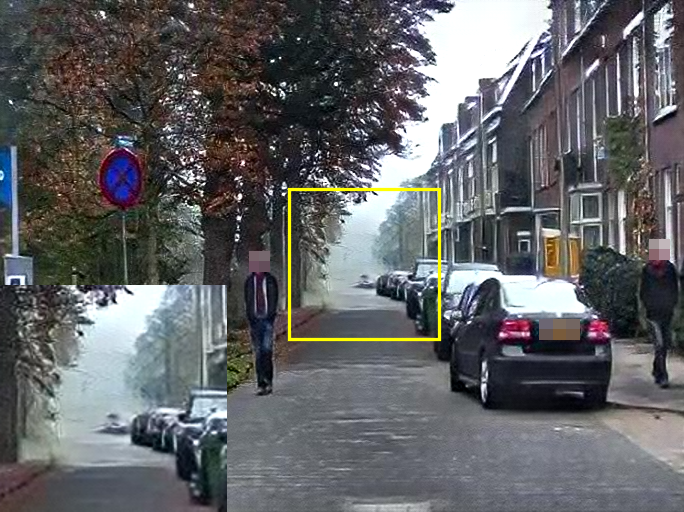}}\hfill
	\subfloat[Dehamer'22~\cite{guo2022image}]{\includegraphics[width = 0.196\textwidth]{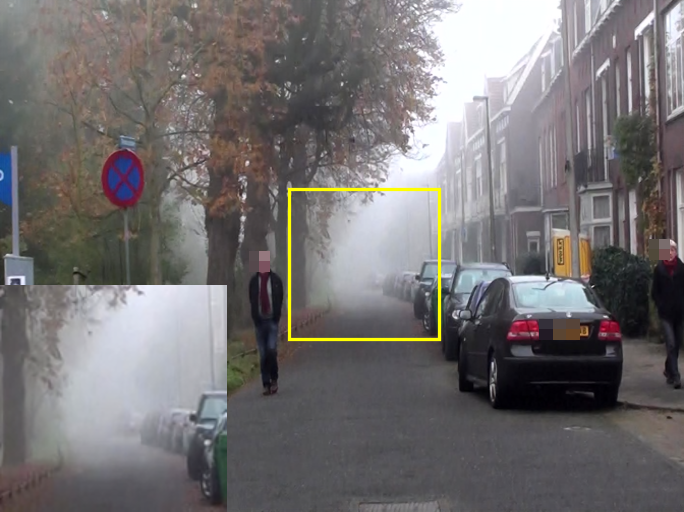}}\hfill
	\subfloat[DehazeF.'22~\cite{song2022vision}]{\includegraphics[width = 0.196\textwidth]{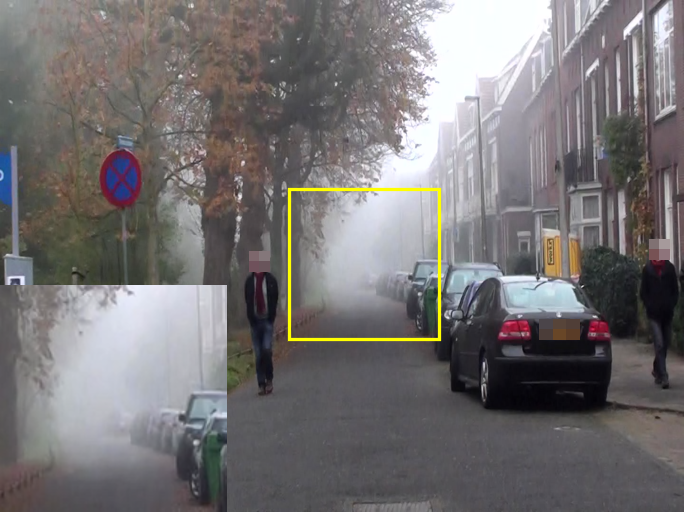}}\hfill
	\subfloat[D4'22~\cite{yang2022self}]{\includegraphics[width = 0.196\textwidth]{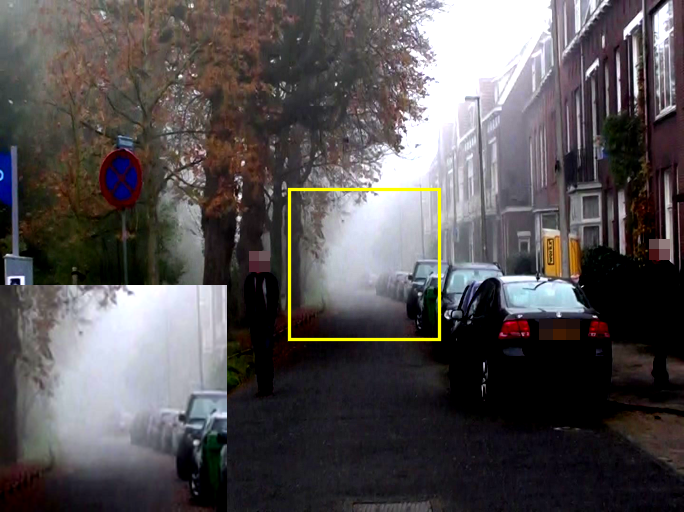}}\hfill
	\subfloat[PSD'21~\cite{chen2021psd}]{\includegraphics[width = 0.196\textwidth]{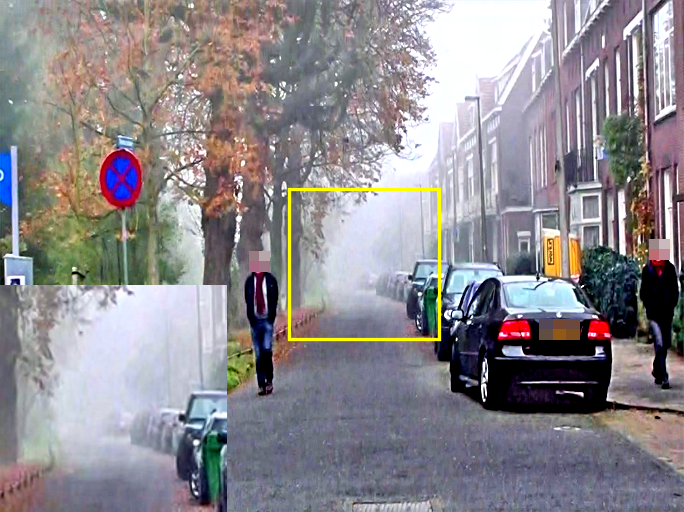}}\hfill
	\subfloat[4K'21~\cite{zheng2021ultra}]{\includegraphics[width = 0.196\textwidth]{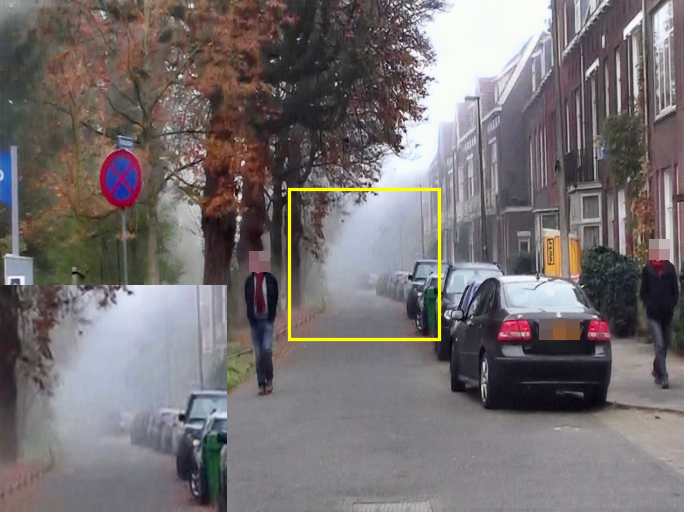}}\hfill
	\subfloat[MSBDN~\cite{dong2020multi}]{\includegraphics[width = 0.196\textwidth]{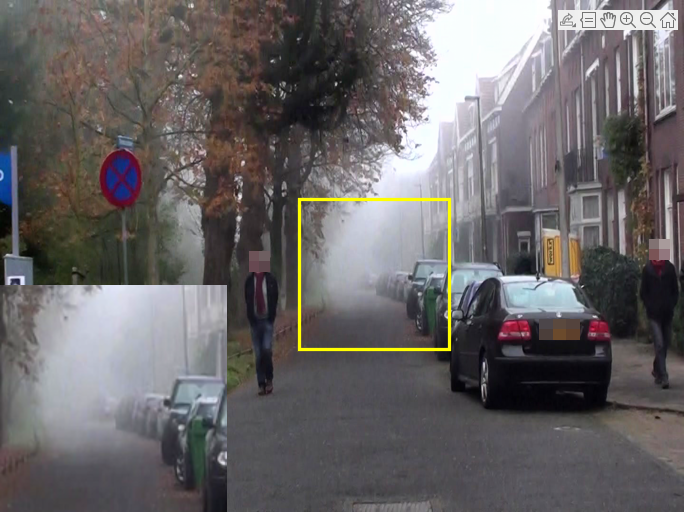}}\hfill
	\subfloat[DAN~\cite{shao2020domain}]{\includegraphics[width = 0.196\textwidth]{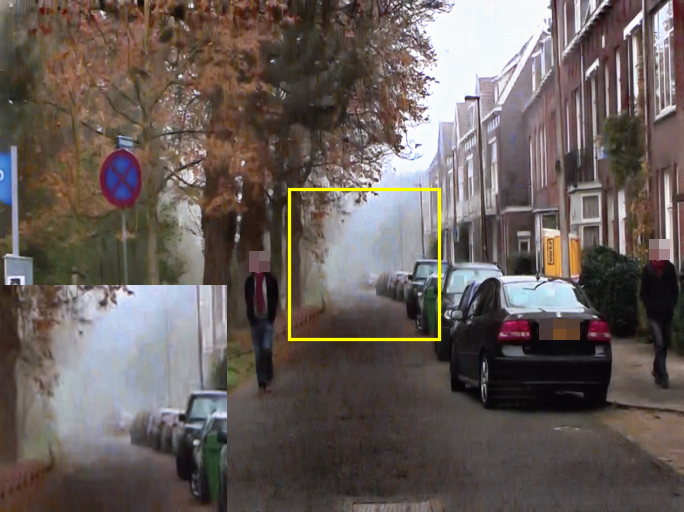}}\hfill
	\subfloat[GDN~\cite{liu2019griddehazenet}]{\includegraphics[width = 0.196\textwidth]{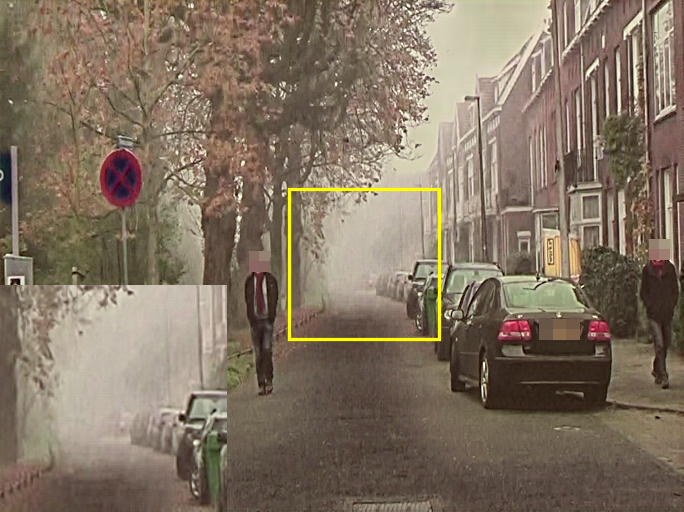}}\hfill
	\setcounter{subfigure}{0}
	\subfloat[Input]{\includegraphics[width = 0.196\textwidth]{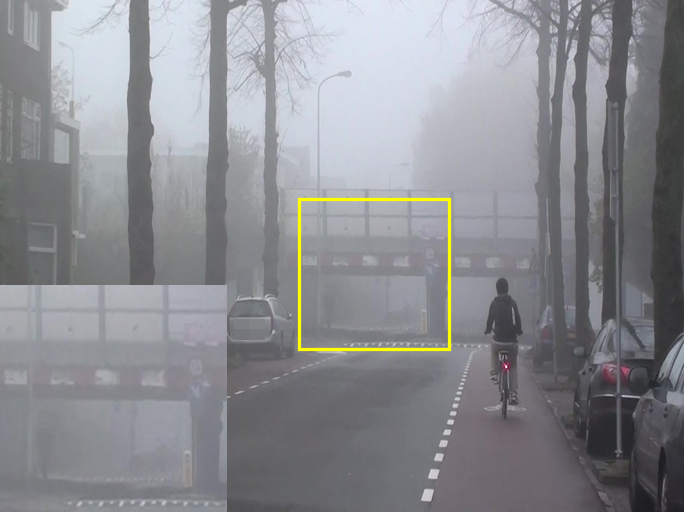}}\hfill
	\subfloat[Ours]{\includegraphics[width = 0.196\textwidth]{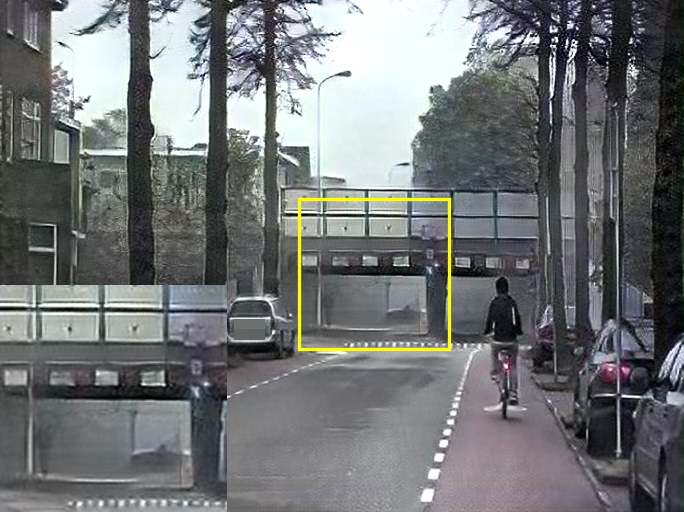}}\hfill
	\subfloat[Dehamer'22~\cite{guo2022image}]{\includegraphics[width = 0.196\textwidth]{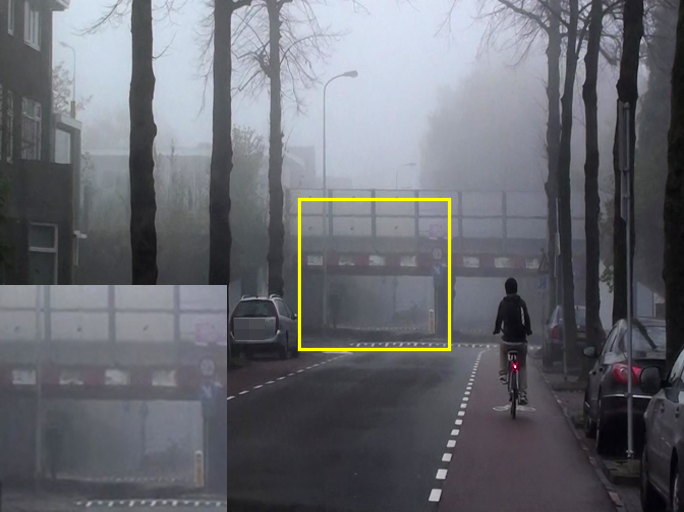}}\hfill
	\subfloat[DehazeF.'22~\cite{song2022vision}]{\includegraphics[width = 0.196\textwidth]{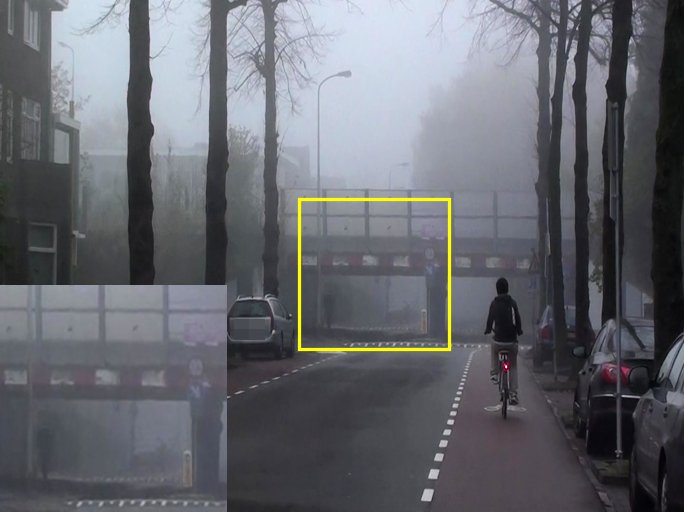}}\hfill
	\subfloat[D4'22~\cite{yang2022self}]{\includegraphics[width = 0.196\textwidth]{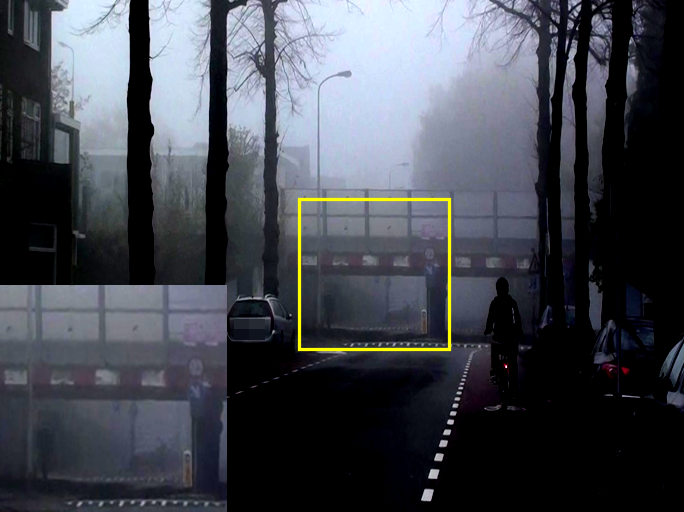}}\hfill
	\subfloat[PSD'21~\cite{chen2021psd}]{\includegraphics[width = 0.196\textwidth]{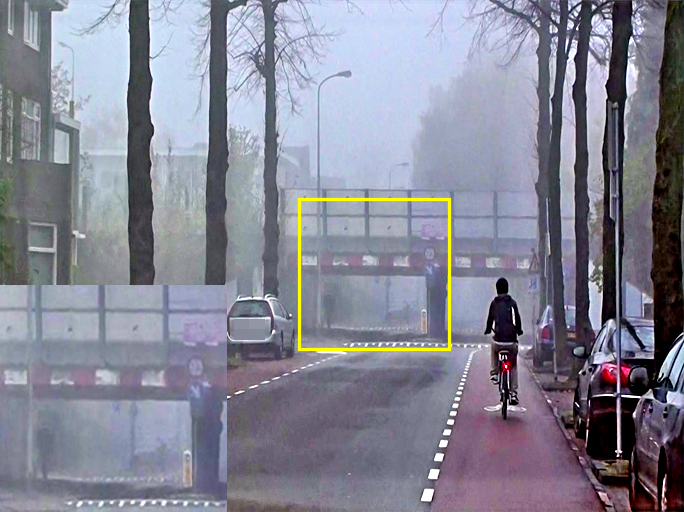}}\hfill
	\subfloat[4K'21~\cite{zheng2021ultra}]{\includegraphics[width = 0.196\textwidth]{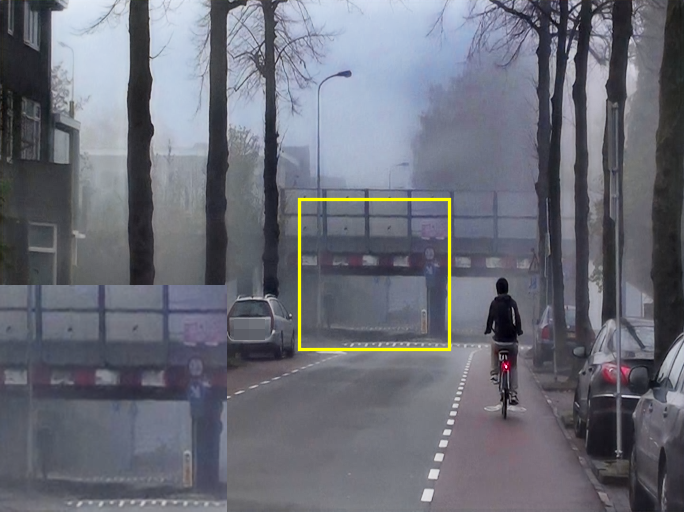}}\hfill
	\subfloat[MSBDN~\cite{dong2020multi}]{\includegraphics[width = 0.196\textwidth]{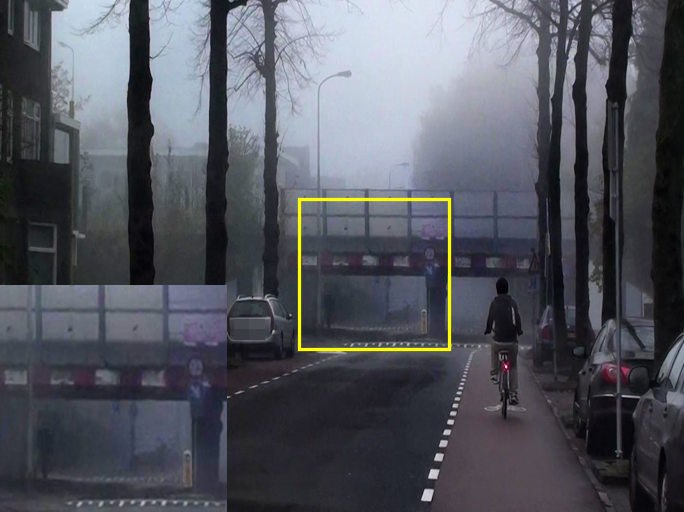}}\hfill
	\subfloat[DAN~\cite{shao2020domain}]{\includegraphics[width = 0.196\textwidth]{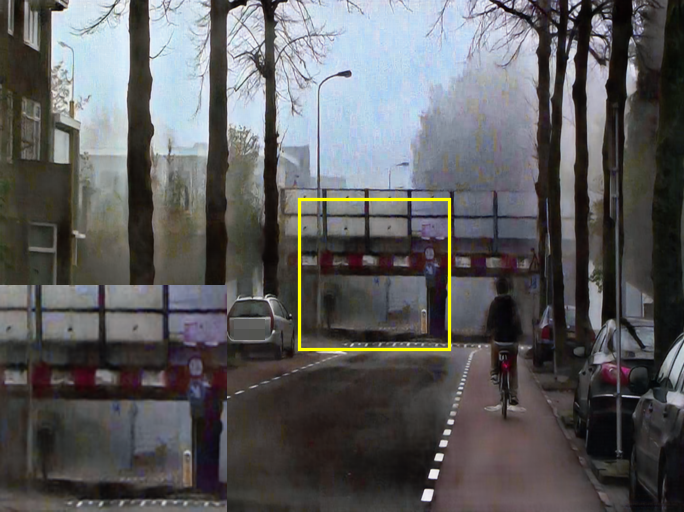}}\hfill
	\subfloat[GDN~\cite{liu2019griddehazenet}]{\includegraphics[width = 0.196\textwidth]{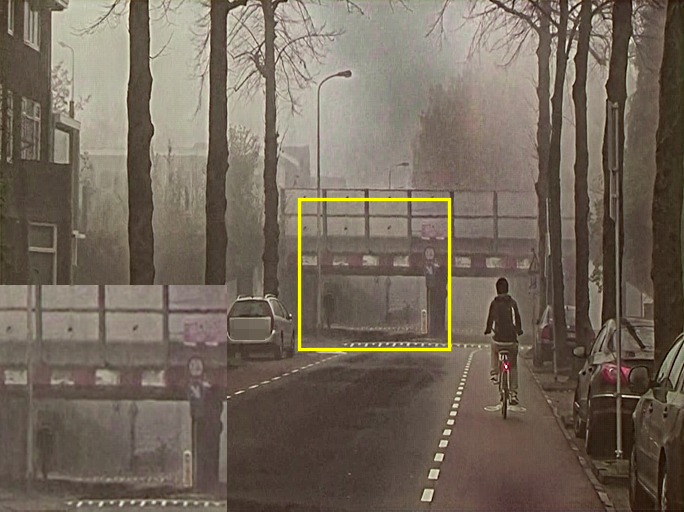}}\hfill
	\caption{Comparison results on real dense fog images. (a) Input images. (b) Our results. (c)$\sim$(g) Results of the state-of-the-art methods. Our results show better visibility.}
	\label{fig:real2}
\end{figure*}

\subsubsection{Baselines}
We evaluate our method against the non-learning method Non-Local Image Dehazing (NLD)~\cite{berman2018single}, state-of-the-art transformer-based dehazing methods~\cite{guo2022image,song2022vision}, CNN-based methods: GridDehazeNet (GDN)~\cite{liu2019griddehazenet}, Domain Adaptation Network (DAN)~\cite{shao2020domain}, Multi-Scale Boosted Dehazing Network (MSBDN)~\cite{dong2020multi}, 4KDehazing (CVPR21)~\cite{zheng2021ultra}, PSD (CVPR21)~\cite{chen2021psd}, D4 (CVPR22)~\cite{yang2022self}, etc.

\subsubsection{Qualitative Comparisons}
Comparisons on the self-collected fog and O-HAZE dataset are shown in Fig.~\ref{fig:ohaze}.
The baseline methods do not perform well on the images. Some results are too dark, and some still have fog left. 
Also, since the generated fog is not uniform in the Dense-Haze and NH-Haze datasets, some fog still remains. 
The deep learning baselines are not able to defog such dense fog adequately.

Figs.~\ref{fig:real1} to~\ref{fig:real2} show the input dense non-uniform fog images, our defogging results, and the results of the state-of-the-art methods.
Due to the uniform and/or severe fog density, the input images are degraded by multiple factors like blur, contrast, sharpness, and color distortion. 
As shown in the figures, our method outperforms the state-of-the-art methods on real dense fog images.

\subsubsection{Quantitative Comparisons}
We also conduct quantitative evaluations on O-HAZE, NH-Haze and Dense-Haze, which are shown in Table~\ref{tab:comparisons}. 
We measure the restoration performance using the Peak Signal-to-Noise Ratio (PSNR) and the Structural Similarity (SSIM); higher is better.
Our method achieves the best PSNR, SSIM performance.

\subsection{Ablation Studies}
\begin{figure}[t!]
	\centering
	\captionsetup[subfloat]{font=small, labelformat=empty}
	\captionsetup[subfloat]{farskip=1pt}
	\setcounter{subfigure}{0}
	\subfloat[Input]{\includegraphics[width = 0.196\columnwidth]{fig/trailer/in.png}}\hfill
	\subfloat[w/o Gray $\mathcal{L}_{\rm m}$\label{fig:abgray}]{\includegraphics[width = 0.196\columnwidth]{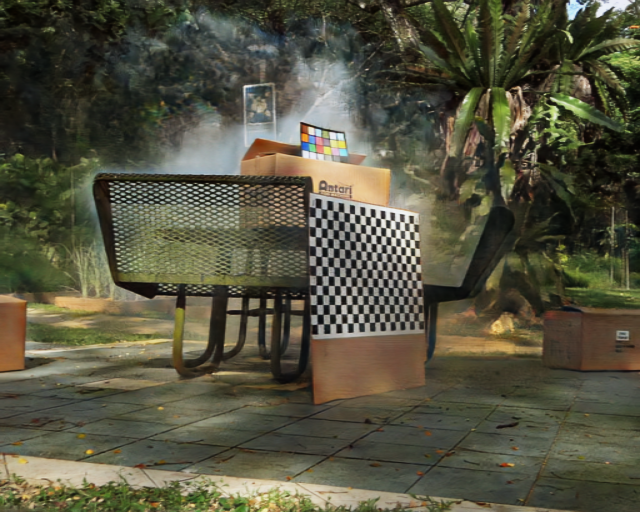}}\hfill
	\subfloat[ViT $S(\mathbf{\hat{J}_Y})$]{\includegraphics[width = 0.196\columnwidth]{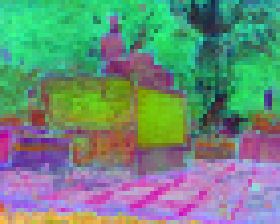}}\hfill
	\subfloat[w/o ViT $\mathcal{L}_{\rm s}$\label{fig:abvit}]{\includegraphics[width = 0.196\columnwidth]{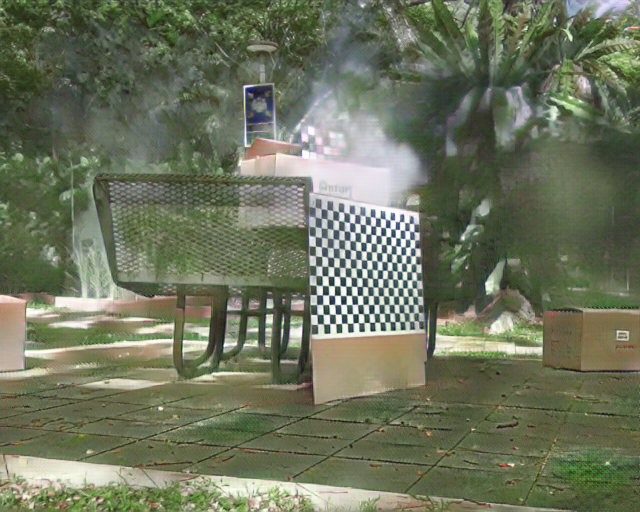}}\hfill
	\subfloat[Output\label{fig:o}]{\includegraphics[width = 0.196\columnwidth]{fig/trailer/our.png}}\hfill
	\caption{Ablation studies on (b) without using our grayscale multiplier consistency loss $\mathcal{L}_{\rm multiplier}$, and (d) without using our Dino-ViT structure consistency loss $\mathcal{L}_{\rm structure}$. (e) is our final output. (c) shows DINO-ViT capture scene structure, helping the network to recover the background information.}
	\label{fig:ablation_vit}
\end{figure}

\begin{figure}[t!]
	\captionsetup[subfigure]{font=small, labelformat=empty}
	\captionsetup[subfloat]{farskip=1pt}
	\centering
	\setcounter{subfigure}{0}
	\subfloat[Input]{\includegraphics[width = 0.196\columnwidth]{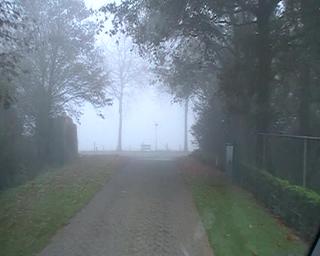}}\hfill
	\subfloat[w/o feedback\label{fig:abfeed}]{\includegraphics[width = 0.196\columnwidth]{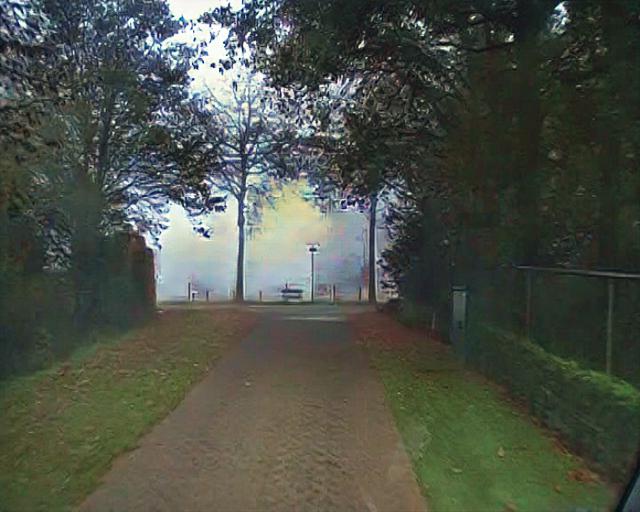}}\hfill
	\subfloat[w/o $\mathcal{L}_{\rm m}$\label{fig:abmultiplier}]{\includegraphics[width = 0.196\columnwidth]{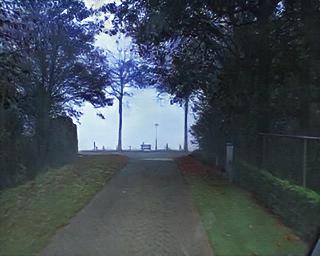}}\hfill
	\subfloat[w/ $\mathcal{L}_{\rm MSE}$ only\label{fig:abmse}]{\includegraphics[width = 0.196\columnwidth]{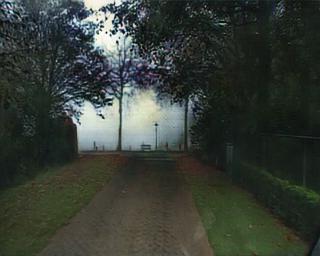}}\hfill
	\subfloat[Output]{\includegraphics[width = 0.196\columnwidth]{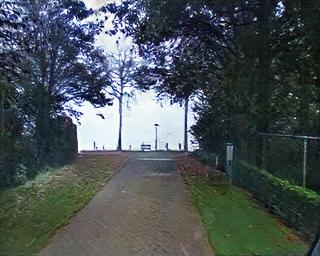}}\hfill
	\caption{Ablation studies on
		(b) without uncertainty feedback network; 
		(c) without multiplier consistency loss; 
		(d) defogging results from our model with the MSE loss only.
		(e) is our final output. }
	\label{fig:ablation}
\end{figure}

We conduct ablation studies to analyze the characteristics of the proposed algorithm.
We first evaluate the grayscale feature multiplier, if the grayscale network is removed, the RGB network will have no guidance from grayscale and the results are shown in Fig.~\ref{fig:abgray}.
To show the effectiveness of using ViT, we remove the structure consistency loss, the results are shown in Fig.~\ref{fig:abvit}. 
We then remove the uncertainty feedback network from our model. After training with the same semi-supervised training strategy and the same loss functions, the results are shown in Fig.~\ref{fig:abfeed}. 
We can observe that the results are not as effective as those using the feedback network. 
In Fig.~\ref{fig:abmultiplier}, we replace our multiplier generator with the normal generator. 
Therefore, we can observe more fake content, such as the fake leaves on the tree. 
Finally, Fig.~\ref{fig:abmse} shows the results of using the MSE loss only. 
The typical results of fully supervised deep learning methods trained on synthetic images are unsatisfactory. 
Some fog still remains, details are lost, and some regions are considerably dark.

\section{Conclusion}
\label{sec:outro}
We have proposed a learning-based defogging method that targets dense and/or non-uniform fog.
Our method combines the structure representations from ViT and the features from CNN as feature regularization that can guide our network to recover background information.
Our pipeline consists of a grayscale network and an RGB network. 
We introduced the grayscale feature multiplier, which is designed to enhance features.
Aside from the new structure loss and multiplier consistency loss, we also introduced uncertainty feedback learning that refines the performance of the RGB generator network. 
Experimental results show that our method works for dense and/or non-uniform fog, and outperforms the state-of-the-art methods.

\section*{Acknowledgment}
This research/project is supported by the National Research Foundation, Singapore under its AI Singapore Programme (AISG Award No: AISG2-PhD/2022-01-037[T]), and partially supported by MOE2019-T2-1-130. Wenhan Yang's research is supported by Wallenberg-NTU Presidential Postdoctoral Fellowship. Robby T. Tan's work is supported by MOE2019-T2-1-130.

\bibliographystyle{splncs}
\bibliography{egbib}
\end{document}